%% file: main.tex
\documentclass[10pt,twocolumn,letterpaper]{article}

\usepackage[pagenumbers]{cvpr} % To force page numbers, e.g. for an arXiv version

\input{preamble}

\usepackage{booktabs}
\usepackage{bm}
\usepackage{makecell}
\usepackage{enumitem}
\usepackage{multirow}
\makeatletter
\@namedef{ver@everyshi.sty}{}
\makeatother
\usepackage{booktabs}
\usepackage{multirow}
\usepackage[normalem]{ulem}
\usepackage{ifsym}

\usepackage{tikz}
\usepackage{comment}

\usepackage[accsupp]{axessibility}  % Improves PDF readability for those with disabilities.

\usepackage{color}
\usepackage{contour}
\usepackage{ulem}
\usepackage{textcomp}
\usepackage{soul}
\usepackage{wrapfig}

\contourlength{0.5pt}

\newcommand\blfootnote[1]{%
\begingroup
\renewcommand\thefootnote{}\footnote{#1}%
\addtocounter{footnote}{-1}%
\endgroup
}

\definecolor{cvprblue}{rgb}{0.21,0.49,0.74}
\usepackage[pagebackref,breaklinks,colorlinks,citecolor=cvprblue]{hyperref}

\def\paperID{5} % *** Enter the Paper ID here % unlucky numbers

\def\pname{Re-Nerfing}

\title{Re-Nerfing: Improving Novel View Synthesis through Novel View Synthesis}

\author{
\quad Felix Tristram$^{*,1}$
\quad Stefano Gasperini$^{*,1,2}$
\quad Nassir Navab$^{1}$
\quad Federico Tombari$^{1,3}$\\
$^1$ Technical University Munich \quad $^2$ VisualAIs \quad $^3$ Google
}

\begin{document}
\maketitle
\blfootnote{$^{*}$ The authors contributed equally.}
\blfootnote{Contact author: Felix Tristram (\textit{felix.tristram@tum.de}).}

\input{sec/0_abstract}    
\input{sec/1_intro}
\input{sec/2_relatedwork}
\input{sec/3_method}

\input{sec/4_experiments}
\input{sec/5_conclusion}

\setcounter{figure}{3}
\setcounter{table}{5}

\input{sec/X_suppl}
{

    \clearpage
    \small
    \bibliographystyle{ieeenat_fullname}
    \bibliography{main}
}

\end{document}

%% file: preamble.tex
\usepackage[dvipsnames]{xcolor}

%% file: sec/0_abstract.tex
\begin{abstract}
Recent neural rendering and reconstruction techniques, such as NeRFs or Gaussian Splatting, have shown remarkable novel view synthesis capabilities but require hundreds of images of the scene from diverse viewpoints to render high-quality novel views. With fewer images available, these methods start to fail since they can no longer correctly triangulate the underlying 3D geometry and converge to a non-optimal solution. These failures can manifest as floaters or blurry renderings in sparsely observed areas of the scene. 
In this paper, we propose \pname, a simple and general add-on approach that leverages novel view synthesis itself to tackle this problem. 
Using an already trained NVS method, we render novel views between existing ones and augment the training data to optimize a second model. This introduces additional multi-view constraints and allows the second model to converge to a better solution. 
%Under the assumption that some scene geometry could be reconstructed during the initial optimization of the given novel view synthesis method, we render previously unseen viewpoints of the scene and use them as additional supervision signals for consecutive optimizations. This introduces additional multi-view constraints and allows further optimization stages to converge to better solutions. 
%In this paper we propose \pname, a simple and general multi-stage data augmentation approach that leverages novel view synthesis to address these limitations. With \pname, we enhance the geometric consistency and thus novel views as follows: First, we train a NVS method with the available images. With this optimized method we then render novel viewpoints (pseudo-views) around the original ones with a view selection strategy to improve the coverage of the scene while maintaining high quality renderings. Finally, we train the same NVS method again from scratch but this time also giving it the pseudo-views as input in addition to the original images. 
With \pname\ we achieve significant improvements upon multiple pipelines based on NeRF and Gaussian-Splatting in sparse view settings of the mip-NeRF 360 and LLFF datasets. Notably, \pname\ does not require prior knowledge or extra supervision signals, making it a flexible and practical add-on. 
%To showcase the efficacy of \pname\, we present extensive experiments applying it on various pipelines on the mip-NeRF 360 and LLFF datasets in sparse view settings and achieve substantial improvements, notably without adding any external data or supervision.
%Neural Radiance Fields (NeRFs) have shown remarkable novel view synthesis capabilities even in large-scale, unbounded scenes, albeit requiring hundreds of views or introducing artifacts in sparser settings. Their optimization suffers from shape-radiance ambiguities wherever only a small visual overlap is available. This leads to erroneous scene geometry and artifacts. In this paper, we propose \pname, a simple and general multi-stage data augmentation approach that leverages NeRF's own view synthesis ability to address these limitations. With \pname, we enhance the geometric consistency of novel views as follows: First, we train a NeRF with the available views. Then, we use the optimized NeRF to synthesize pseudo-views around the original ones with a view selection strategy to improve coverage and preserve view quality. Finally, we train a second NeRF with both the original images and the pseudo views masking out uncertain regions. Extensive experiments applying \pname\ on various pipelines on the mip-NeRF 360 dataset, including Gaussian Splatting, provide valuable insights into the improvements achievable without external data or supervision, on denser and sparser input scenarios.
\end{abstract}

%% file: sec/1_intro.tex
\section{Introduction}
\label{sec:intro}

Novel view synthesis (NVS) methods based on Neural Radiance Field (NeRF)~\cite{mildenhall2020nerf} and 3D Gaussian Splatting (3DGS) ~\cite{kerbl2023gaussiansplat}
have recently revolutionized 3D scene representation and rendering, enabling unprecedented quality in synthesizing novel views from a set of images.
%Among these works, Neural Radiance Field (NeRF)~\cite{mildenhall2020nerf} optimizes a continuous volumetric scene function in the weights of two multi-layer perceptrons (MLPs), which, when queried with rays sampled from posed cameras, generate corresponding RGB values and volume densities~\cite{mildenhall2020nerf}. 3D Gaussian Splatting (3DGS) ~\cite{kerbl2023gaussiansplat} represents the scene explicitly with Gaussian-shaped blobs whose position, size, and orientation are optimized to fit the views given during training~\cite{kerbl2023gaussiansplat}. 
These techniques have been applied across various tasks, simplifying content creation, rendering, and reconstruction workflows.

%Therefore, a plethora of works has been proposed to mitigate these issues, leading to faster training times~\cite{iNGP} and techniques to improve the robustness of NeRF~\cite{BARF,etc}.

%\begin{wrapfigure}{l}{0.5\textwidth}
\begin{figure}[t]
\begin{center}
\includegraphics[width=1.0\linewidth]{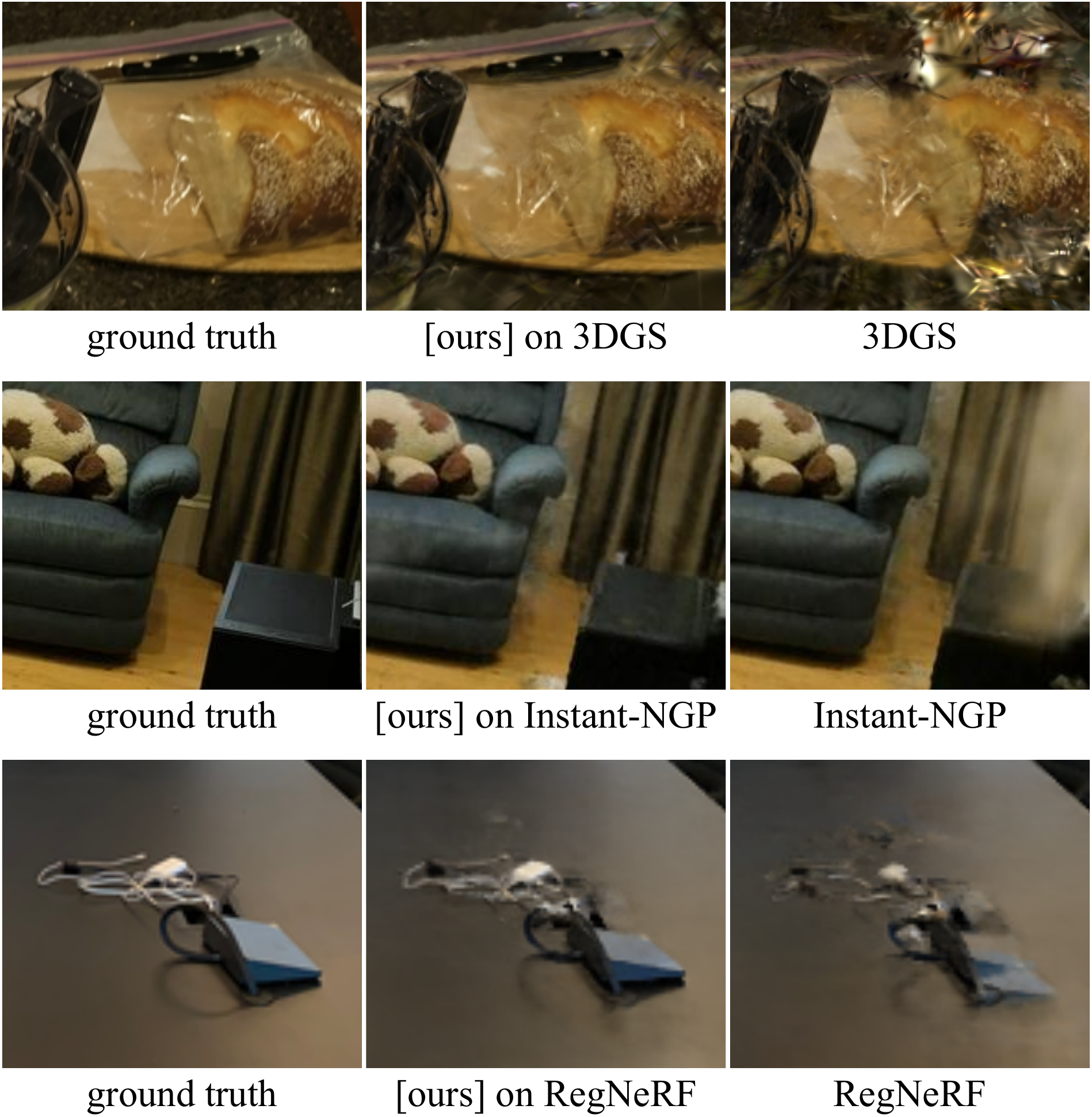}
\vspace{-0.7cm}
\end{center}
   \caption{   
   Examples of synthesized views by 3DGS~\cite{kerbl2023gaussiansplat}, Instant-NGP~\cite{muller2022instant}, 
   RegNeRF ~\cite{niemeyer2022regnerf}, with and without the proposed \pname.
   Our approach improves NVS by improving the optimization, reaching better global solutions thanks to the addition of novel views synthesized by the baseline model to the training data.
   Image crops show test views from the mip-NeRF 360~\cite{barron2022mip360} dataset for 3DGS, iNGP trained on 30 views and from the LLFF~\cite{mildenhall2019llff} dataset for RegNeRF trained on 3 views.}
\label{fig:teaser}
\vspace{-0.25cm}
\end{figure}
%\end{wrapfigure}

Despite its remarkable success, NeRF~\cite{mildenhall2020nerf} has several limitations, such as slow training~\cite{muller2022instant}, failures when noisy or no camera poses are available~\cite{lin2021barf,bian2023nope}, and computationally intense rendering mostly incompatible with mobile applications~\cite{chen2023mobilenerf}. Many works have been proposed to address these issues, such as 3DGS~\cite{kerbl2023gaussiansplat} changing representation to shorten training times and simplifying inference. However, collecting enough high-quality images with sufficient overlap to ensure successful model convergence remains a challenging task. Towards this end, researchers have explored the application of NVS in sparse view settings, with only a handful of images available~\cite{truong2023sparf,niemeyer2022regnerf,zhu2023fsgs}.

Complex geometries and large-scale environments often lead to artifacts due to the inability to correctly triangulate the scenes' 3D structure ~\cite{roessle2022dense,barron2021mip,deng2022dsnerf}. This is particularly severe in sparse settings, where the limited views lead to hallucinations, e.g. floaters in  free space due to the ambiguity of shape and radiance~\cite{truong2023sparf,deng2022dsnerf,martin2021nerfwild}.
To mitigate these issues, depth and structural priors have been introduced to the optimization to leverage additional geometric information~\cite{deng2022dsnerf,roessle2022dense,truong2023sparf}. However, reliable dense depth estimates are also hard to obtain, and using wrongly estimated depth can further hurt performance~\cite{deng2022dsnerf}.
% add sentence about how many depend on external methods and data........

In this work, we mitigate these open issues with a simple and effective add-on solution exploiting the inherent novel view synthesis capability of NVS methods. We achieve this with \pname, a multi-stage, general pipeline that introduces structural constraints from synthetic views generated by a previously trained model on the same scene. As shown in Figure~\ref{fig:teaser}, thanks to its ability to improve the view coverage while preserving quality discarding uncertain regions, \pname\ significantly improves the synthesis of new views.
The main contributions of this paper can be summarized as follows:
\begin{itemize}
    \item With \pname, we show how NeRF's and Gaussian Splatting's novel view synthesis is beneficial to enhance their own rendering quality.
    \item We propose a simple and effective iterative augmentation technique to enhance any NVS outputs by training a new NVS model with the addition of synthesized views selected to improve the scene coverage and masked to discard uncertain areas.
    %\item We introduce a novel density loss for NeRF optimization derived from epipolar geometry that is generally applicable to any stereo-setup to be used with NeRF.
    %\item We provide insights on how the views should be selected and show how masking their uncertain regions can help mitigating a sub-optimal selection.
    \item We show the wide applicability of \pname\ upon various NVS pipelines~\cite{kerbl2023gaussiansplat,turki2024pynerf,muller2022instant,niemeyer2022regnerf} and datasets~\cite{barron2022mip360,mildenhall2019llff}.
    %improving in both denser and sparser settings.
    %\item We propose to evaluate the synthesized view regions according to their visibility within the training views.
\end{itemize}
%We apply \pname\ on top of various pipelines~\cite{tancik2023nerfstudio,turki2024pynerf,muller2022instant} and improve in both denser and sparser settings.

%% file: sec/2_relatedwork.tex
\section{Related Work}
\label{sec:relwork}
%\pname{} constitutes a method that brings together the fields of NVS and stereo based disparity estimation which we both consecutively review.

%\subsection{View Synthesis and Neural Representations}
Given a dense set of images, traditional interpolation techniques based on light field sampling can synthesize photorealistic novel views~\cite{levoy1996lightfield}. In relatively sparser settings, neural networks have been used for view synthesis by exploiting the correlation of the visual appearance of an instance across different views~\cite{zhou2016appearance_flow,flynn2016deepstereo}. More recently, 3DGS~\cite{kerbl2023gaussiansplat} has shown great promise in view synthesis tasks by optimizing the scene as 3D Gaussians with spherical harmonics.

Neural implicit representations have emerged as a shift from traditional explicit representations like point clouds, meshes, and voxels~\cite{qi2017pointnet,zhou2018voxelnet}, by encoding 3D shapes implicitly in the weights of a neural network. The pioneering works of Occupancy Networks~\cite{Mescheder2019occupancy} and DeepSDF~\cite{Park2019deepsdf} initiated this line of work by encoding complex shapes in a continuous function space through a network. Unlike traditional representations, here, no space discretization is needed, enabling finer details~\cite{sitzmann2019scene}.

\textbf{NeRF}
Mildenhall et al.~\cite{mildenhall2020nerf} introduced NeRF by combining neural implicit representations with differentiable rendering, encoding both volumetric properties and view-dependent appearance and leading to high-quality novel views. NeRF has catalyzed a myriad of works tackling various aspects of its limitations and further improving NVS~\cite{xie2022nerf_survey}. Among these, Instant-NGP~\cite{muller2022instant}, FastNeRF~\cite{garbin2021fastnerf} and others~\cite{tensorf, SunSC22, yu_and_fridovichkeil2021plenoxels} enable significantly faster training speeds, BARF~\cite{lin2021barf}, SPARF~\cite{truong2023sparf} and Nope-NeRF~\cite{bian2023nope} cope with imperfect or absent camera poses, mip-NeRF 360~\cite{barron2022mip360} deals with unbounded scenes and Nerfies~\cite{park2021nerfies} among others ~\cite{Cao2022FWD, park2021hypernerf, kplanes_2023} deals with dynamic scenes. Additionally, the work of R{\"o}ssle et al.~\cite{roessle2022dense} delivers more consistent geometry, CoNeRF~\cite{kania2022conerf} enables altering the output, Panoptic Neural Fields~\cite{kundu2022panopticnerf} incorporates scene understanding information, Block-NeRF~\cite{tancik2022blocknerf} renders city-scale scenes, and CamP~\cite{park2023camp} optimizes the camera parameters. 
PyNeRF~\cite{turki2024pynerf} combines the benefits of mip-NeRF~\cite{barron2021mip} and grid-based models, such as Instant-NGP, to facilitate fast training and rendering of anti-aliased images. 
%Zip-NeRF~\cite{barron2023zip} combines mip-NeRF 360 and grid-based models such as Instant-NGP to reach state-of-the-art results with fast training speeds. 
%3D Gaussian Splatting~\cite{kerbl2023gaussiansplat} exploits the efficiency of explicit methods and represents the scene with 3D Gaussians.

\textbf{Sparse settings}
Nevertheless, highly sparse settings are particularly challenging as the standard multi-view constraints are not sufficient~\cite{truong2023sparf,choi2019extreme}.
Strong geometric information is required to reconstruct the scene and enable satisfactory novel views synthesis. Towards this end, researchers have explored the support of additional data and external models to provide geometric supervision~\cite{truong2023sparf,roessle2022dense} or semantic knowledge~\cite{yu2021pixelnerf,rebain2022lolnerf}. Such techniques enable NVS even from a single image.
SPARF~\cite{truong2023sparf}, for example, relies on a pixel correspondence network to enforce a geometrically accurate solution. PixelNeRF~\cite{yu2021pixelnerf} learns a prior from multiple scenes and conditions the novel views on the few inputs available. 
For Gaussian Splatting-based methods, multiple extensions have been proposed to deal with sparse view settings~\cite{zhu2023fsgs, li2024dngaussian, paliwal2024coherentgs, fan2024instantsplat}.
However, all aforementioned methods depend on external models and large amounts of training data, which eliminates the advantage of NeRF and 3DGS of not requiring any training data other than that for the object or scene of interest.

\textbf{Geometric constraints}
Enforcing geometric consistency plays a crucial role in NVS as it directly impacts the quality of the generated views~\cite{roessle2022dense}. DS-NeRF~\cite{deng2022dsnerf} supervises NeRF's optimization with sparse depth obtained from COLMAP~\cite{schonberger2016colmap}, thereby reducing overfitting and accelerating training. R{\"o}ssle et al.~\cite{roessle2022dense} went a step further by feeding the sparse COLMAP point cloud to a depth completion network and constraining NeRF's optimization according to the estimated depth and its associated uncertainty. Urban Radiance Fields~\cite{rematas2022urban} focuses on outdoor scenes where LiDAR data is available, which is used for losses that benefit surface estimation.
FSGS~\cite{zhu2023fsgs} and related methods~\cite{paliwal2024coherentgs, li2024dngaussian} on the other hand rely on monocular depth prediction to acquire a good geometry initialisation and supervision signal. 
%VoxNeRF~\cite{wang2023voxnerf} leverages a volumetric representation as a sparse voxel octree within a probabilistic framework to account for the uncertainty of the surface-ray intersection point. 
SPARF~\cite{truong2023sparf} exploits multi-view geometry constraints and pixel correspondences, minimizing the re-projection error using the depth rendered by the NeRF and the pose estimates, which are optimized jointly.
%Epipolar geometry is a well-established concept in computer vision describing the geometric relationship between views captured from different perspectives. It provides constraints on the location of corresponding points in different images~\cite{hartley2003multipleviewgeom}. These principles have been applied for 3D reconstruction~\cite{labatut2007efficient} and are fundamental for stereo depth estimation~\cite{kanade1994stereo}.

\begin{figure*}[t]
\begin{center}
\includegraphics[width=1.00\textwidth]{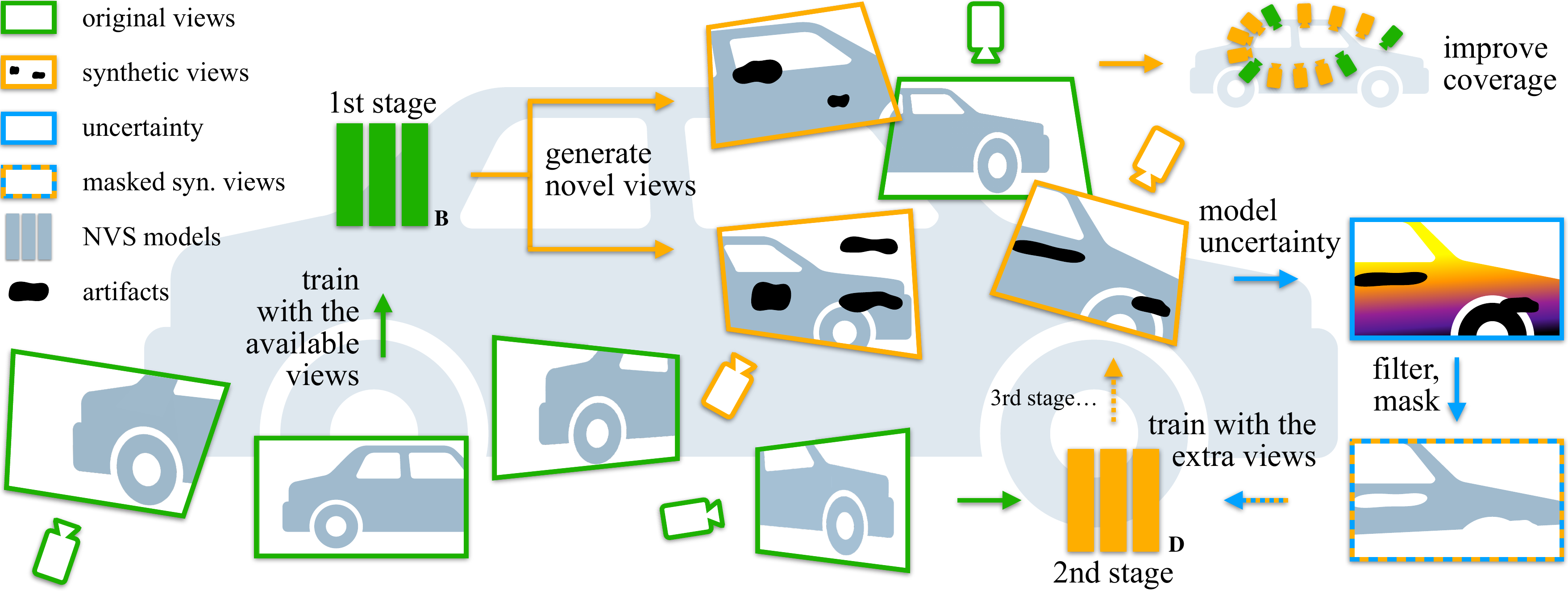}
\vspace{-0.7cm}
\end{center}
   \caption{\pname\ is a multi-stage framework. Compatible with any NVS pipeline, it operates by first training a model with the available views (green, 1st). This model is then used to generate novel views from camera poses to improve the scene coverage (orange). Then, we compute the model's uncertainty on such novel views (blue) and discard the uncertain regions (orange-blue). Finally, we use these masked views and the original ones to train a second model (orange, 2nd). The process can be repeated iteratively by generating the novel views with the second model (3rd stage).}
\label{fig:framework}
\vspace{-0.25cm}
\end{figure*}

\textbf{Data augmentation for NVS}
In general, data augmentation aims to increase the pool of training data with samples that are variations of the existing ones (e.g., image flip), ultimately regularizing the model.
As for other computer vision tasks and settings~\cite{nekrasov2021mix3d,ghiasi2021simple,lehner2023ijcv}, data augmentation techniques have also been proposed for NVS. In this domain, PANeRF~\cite{ahn2022panerf}, GeoAug~\cite{chen2022geoaug}, and VM-NeRF~\cite{bortolon2023vmnerf} generate new views by warping existing ones via homography~\cite{ahn2022panerf} or exploiting the NeRF's depth estimates~\cite{chen2022geoaug,bortolon2023vmnerf}. In contrast, Aug-NeRF~\cite{chen2022augnerf} trains more robust NeRFs with worst-case perturbations of the input coordinates, the intermediate features, and the pre-rendering outputs.

\textbf{Distillation and semi-supervised learning}
Knowledge distillation involves transferring knowledge from a large, complex model (teacher) to a smaller, more efficient one (student)~\cite{hinton2015knowledgedistillation}. This paradigm has been widely adopted and extended to semi-supervised learning, from segmentation~\cite{chen2020naivestudent} to depth estimation~\cite{gasperini2023md4all}. Naive-student~\cite{chen2020naivestudent} predicts pseudo-labels for a set of unlabeled data and later trains a new model using the original data and the newly pseudo-labeled samples. Tosi et al.~\cite{tosi2023nerfstereo} achieved remarkable depth estimates by training a stereo depth model using data synthesized by NeRF.
NeRFmentation~\cite{feldmann2024nerfmentation} trains a NeRF model and uses it to generate augmented data for a monocular depth estimator.

In this work, we enhance NVS in sparser settings by exploiting the NVS method's inherent view synthesis capabilities. Unlike existing approaches~\cite{truong2023sparf,roessle2022dense,yu2021pixelnerf,rematas2022urban}, we do so in a data augmentation manner by using only the available views without extra data or additional models. We achieve this by training a baseline NVS method on the available data and using it to generate novel views that we add to the training data of a subsequent NVS model. Specifically, we sample such views to improve the scene coverage while preserving the images' quality by masking out uncertain regions.
In contrast to Tosi et al.~\cite{tosi2023nerfstereo} and Feldmann et al.~\cite{feldmann2024nerfmentation}, who trained a depth estimator with NeRF data, we train an NVS model with other NVS data. Then, unlike other augmentation methods~\cite{ahn2022panerf,bortolon2023vmnerf,chen2022geoaug}, our newly introduced views are fully synthesized from an NVS model. 
%As for standard (e.g., flipping and cropping) and more sophisticated augmentations~\cite{ghiasi2021simple,nekrasov2021mix3d,lehner2023ijcv}, which can be mixed to further improve the models~\cite{lehner2023ijcv}, ours can be combined with other augmentation techniques and is not an alternative to them.

%% file: sec/3_method.tex
\section{Method}
\label{sec:method}
As shown in Figure~\ref{fig:framework}, our proposed \pname\ operates in a multi-stage fashion. First, a baseline method is trained with the available views (Section~\ref{sec:first_round}), then it is used to generate novel views to improve the scene coverage (Section~\ref{sec:generate_views}), and lastly, a new model is trained on the original and the synthetic views (Section~\ref{sec:second_round}), discarding uncertain regions of the rendered views to improve the signal's quality.
\pname\ is a general framework that is compatible with any pipeline for NVS. Furthermore, \pname\ follows an iterative process, so further rounds can be executed using the model trained at the previous iteration as a baseline for the next one.

\subsection{NeRF and 3DGS Optimization}
\label{sec:first_round}

To better understand the shortcomings of NeRF~\cite{mildenhall2020nerf} and 3DGS~\cite{kerbl2023gaussiansplat} optimizations in a sparser setting, we review how they optimize their scene representation and render images from it. Both methods are optimized through photometric consistency, where the color of each ray or pixel is compared to the ground truth color with an L2 loss or a mixture of L1 and SSIM losses \cite{kerbl2023gaussiansplat}. 
This color is obtained through volumetric rendering via the following equations, where density $o$, Transmittance $T$, and color $\mathbf{c}$ are taken along a ray or z-value at intervals $\delta_i$: 
\begin{equation}
    C = \sum^{N}_{i} T_i \alpha_i \mathbf{c}_i,
    \vspace{-0.2cm}
\end{equation}
where
\begin{equation}
    \alpha_i = (1-\text{exp}(-o_i\delta_i)) \;\;\; \text{and}\;\;\; T_i = \prod_{j=1}^{i-1}(1-\alpha_j). 
\end{equation}

This formulation holds for both NeRF-based architectures and neural point-based rendering approaches, such as 3DGS~\cite{kerbl2023gaussiansplat}, where the image formation process is the same but images are rendered through ray-tracing or rasterization.  

\begin{wrapfigure}{l}{0.2\textwidth}
  \centering
    \includegraphics[width=0.2\textwidth]{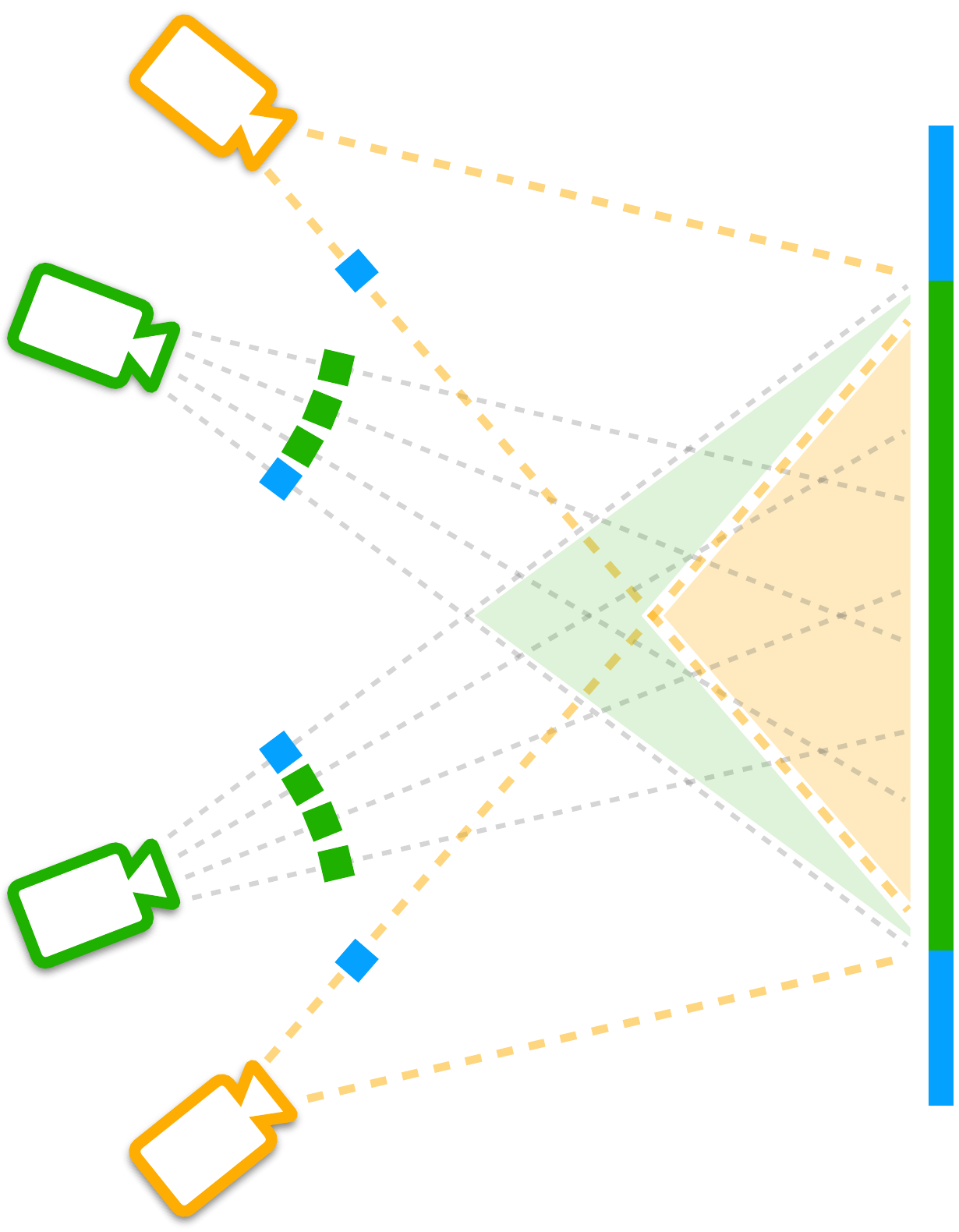}
  %\caption{Birds}
\end{wrapfigure}

To represent the geometry of the given scene only the density $o_i$ at sampling or point locations can be optimized. If very few images of the scene are given, this density can no longer be correctly triangulated since there are not enough multi-view constraints and too much ambiguity in the color supervision. 
This can be seen in the simplified example in the inset, where the green cameras represent the original ones and the shaded green region represents valid solutions where density could be distributed. The overall optimization process would converge to one of the many valid solutions within the green region, i.e., a local minimum, leading to dissatisfying geometry reconstruction and poor NVS. 

\subsection{Augmenting via Synthesized Views}
\label{sec:generate_views}
As introduced above, sparse view reconstruction is a highly under-constrained problem. In scenarios without enough views, e.g., due to crowd-sourced or previously captured images, NVS is particularly difficult. Various methods tackled this with prior knowledge \cite{truong2023sparf, deng2022dsnerf} or regularization \cite{niemeyer2022regnerf}. Instead, we propose to leverage the inherent reconstruction and rendering capabilities of NVS methods to render novel views and utilize these rendered images as \textit{additional training data} in subsequent optimizations (Section~\ref{sec:second_round}), thereby introducing additional multi-view constraints \textit{without} prior knowledge or external models. 

We motivate this approach with the simplified example in the inset figure (Section \ref{sec:first_round}, where the orange cameras represent novel views rendered after the first optimization, and the green cameras are the originally available views. By introducing these additional views, possible solutions reduce from the green-shaded region to the orange-shaded region, making triangulation of the scene's geometry/density easier. We then re-start the optimization from scratch (Section \ref{sec:second_round}). Thanks to the additional viewpoints present in the training data, we can avoid some of the previously present local minima and converge to a better global solution for the given NVS method.

\textbf{View coverage}
We seek to improve the view coverage of the captured scene or object through our augmentations. This is particularly relevant in sparser settings. Toward this end, we generate multiple novel views from camera poses interpolated across existing neighboring views. We define the hyperparameters $N$ and $\Omega$, being the augmentation factor and the set of $N-1$ interpolation factors $\beta_j$, respectively. Specifically, for the simple case of equally spaced novel views:
\begin{equation}
    \Omega = \left\{ \beta_j \mid \beta_j = \frac{j}{N}, \; j = 1, 2, \ldots, N-1 \right\}
\end{equation}

However, equally spaced synthesized views are not ideal as the quality would degrade around $\beta=0.5$. This is particularly relevant in sparser settings where the larger camera displacement between existing poses would lead to worse synthetic views. This is illustrated in the Supplementary Material.

Considering the existing camera poses $\mathbf{P}_i$ and $\mathbf{P}_{i+1}$ with $\mathbf{P} = [\mathbf{p}, \mathbf{q}]$, such that $\mathbf{p}$ is the position vector in 3D space and $\mathbf{q}$ is the quaternion representation of the rotation,
\begin{equation}
%    \mathbf{p}_{\text{interp}} = (1 - \alpha) \mathbf{p}_i + \alpha \mathbf{p}_{i+1}
%    \mathbf{p}_{\text{interp}, j} = (1 - \alpha_j) \mathbf{p}_i + \alpha_j \mathbf{p}_{i+1}
    \bar{\mathbf{p}}_{i,j} = (1 - \beta_j) \mathbf{p}_i + \beta_j \mathbf{p}_{i+1}
\end{equation}
is the interpolation of the position component given an interpolation factor $\beta_j$. For the rotation, we compute the interpolated $\bar{\mathbf{q}}_{i,j}$ from the original pose quaternions $\mathbf{q}_{i}$ and $\mathbf{q}_{i+1}$ using SLERP~\cite{shoemake1985slerp} as follows:
\begin{equation}
%\mathbf{q}_{\text{interp}, j} = \frac{\sin((1 - \alpha_j) \theta)}{\sin(\theta)} \mathbf{q}_i + \frac{\sin(\alpha_j \theta)}{\sin(\theta)} \mathbf{q}_{i+1}
%    \mathbf{q}_{\text{interp}} = \frac{\sin((1 - \alpha) \theta)}{\sin(\theta)} \mathbf{q}_i + \frac{\sin(\alpha \theta)}{\sin(\theta)} \mathbf{q}_{i+1}
\bar{\mathbf{q}}_{i,j} = \frac{\sin((1 - \beta_j) \theta)}{\sin(\theta)} \mathbf{q}_i + \frac{\sin(\beta_j \theta)}{\sin(\theta)} \mathbf{q}_{i+1}
\end{equation}

%half dome and occlusions
Instead of interpolating across existing views, other strategies could generate novel views around a half dome centered at the scene or object center, or following a specific pattern. However, in real-world scenes, such techniques may raise issues with occlusions, as the sampled poses might collide with parts of the scene (e.g., wall) or end up inside an object. Interpolating is more conservative and does not require extra information about the scene, such as the scene occupancy, to avoid collisions and unrealistic viewpoints. Nevertheless, multiple pose sampling strategies could be combined to further enhance the view synthesis.

The formulation above is based on two consecutive poses $\mathbf{P}_i$ and $\mathbf{P}_{i+1}$. In settings where sequential information is not available, a sequence can be simulated by considering consecutive poses from neighboring positions in 3D space~\cite{cover1967nearest}. This can be achieved with a nearest neighbor on the vectors $\mathbf{p}$ using the Euclidean distance or the angular distance between the quaternions $\mathbf{q}$.

\textbf{Image Quality}
Our proposed method has to balance the synthesized image quality with achieving as wide a baseline as possible with regard to the original views. This is because novel views close to the training views will be of higher quality and exhibit less artifacts compared to ones which are further away but add less multi-view constraints. In practice we achieve this trade-off through generating many interpolated views between the original training views $\mathbf{P}_i$ and $\mathbf{P}_{i+1}$ and achieve a good balance between both factors. We illustrate the trade-off in the supplementary material. 

\subsection{\pname: Re-Training NeRF or 3DGS}
\label{sec:second_round}
After training the first model $\mathbf{B}$ and using it to generate novel views (Section~\ref{sec:generate_views}), we train another NVS method $\mathbf{D}$ from scratch with the addition of the synthesized views. To improve the signal's quality, we mask out uncertain regions of the generated views with Bayes' Rays~\cite{goli2023bayes}, where possible.

\textbf{Uncertainty masks}
We seek to add a high-quality training signal to the new model $\mathbf{D}$ by means of the synthesized views. Since the quality of such views is not on par with the originals, they may display artifacts that can impact the renderings. Therefore, we want to remove such artifacts and improve the training signal. 
One way of estimating the reconstruction uncertainty is to use Bayes'Rays~\cite{goli2023bayes}, which can be used to mask out more uncertain regions from our synthesized images. 
%We do so by estimating the reconstruction uncertainty for each synthesized view with Bayes' Rays~\cite{goli2023bayes} and masking out the uncertain regions. 
Bayes' Rays~\cite{goli2023bayes} outputs a volumetric uncertainty field for every input coordinate $\mathbf{x}$:
\begin{equation}
    U(\mathbf{x}) = \text{Trilinear}(\mathbf{x}, \sigma),\;\;\; \text{where} \;\;\; \sigma = ||\bm{\sigma}||_2 
\end{equation}
Here $\bm{\sigma} = (\sigma_x, \sigma_y, \sigma_z)$ represents the diagonal entries of the variance $\bm{\Sigma}$ of the deformation field, which intuitively encodes how much one can permute the underlying NeRF geometry without affecting reconstruction quality. 
$\bm{\Sigma}$ is approximated via the diagonal entries of the Hessian $\bm{H}(\bm{\theta})$, which is approximated for all rays $\bm{r}$ of a sampling pool $R$ with respect to the parameters $\bm{\theta}$ of the deformation field as: 

\begin{equation}
    \bm{H(\theta)} \; \approx \; \frac{2}{R} \; \sum_{\bm{r}} \bm{J_\theta}(\bm{r})^T\bm{J_\theta}(\bm{r}) + 2\lambda\bm{I} 
\end{equation}

\begin{equation}
    \bm{\Sigma} \; \approx \; \text{diag}(\bm{H}(\bm{\theta}))^{-1}
\end{equation}
For the full derivation of this formulation we refer the reader to the original publication of Bayes' Rays~\cite{goli2023bayes}. 

The uncertainty obtained at each sampling location $\mathbf{x}$ can then be rendered via volumetric rendering. 
%The method involves simulating spatially parametrized perturbations of the radiance field and using a Bayesian Laplace approximation to create a volumetric uncertainty field. This field can be rendered similarly to an additional color channel.
Specifically, for each ray $\mathbf{r}$ of a novel pose $\bar{\mathbf{P}}$ with ray samples $\mathbf{x}$, we synthesize the color $C_\mathbf{B}(\mathbf{r})$ and render the corresponding uncertainty $U(\mathbf{r(x)})$. Then, we create pixel-masks where rays with $U(\mathbf{r(x)})$ higher than a threshold $\mu$ are masked out. 
To optimize the second model $\mathbf{D}$, we use the generated views as part of the training images and query their masked rays during the course of optimization. 
%The intuition behind using these synthesized images is that they provide a more diverse yet reliable set of views, which allows the NeRF representation to better triangulate the underlying scene geometry.

\textbf{Training Schedule}
For both NeRF and 3DGS the scene geometry is reconstructed in the early optimization stages and later they are optimized for appearance.
Since the synthesized views are not as good as real images, especially when generated at positions further away from the training views, they can become a sub-optimal signal when optimizing for appearance. 
%Nevertheless, the synthesized views are not as good as real images, especially when synthesized at positions further away from the training views. 
Therefore, we sample from our synthesized images only in the early stages of the optimization, where they help disambiguating the scene geometry, and then remove them from the training as they start to become detrimental to the appearance optimization. 
%We additionally mitigate the quality degradation of large baseline novel views by masking out uncertain regions in the synthesized views (Section~\ref{sec:second_round}).
%This is especially important at the beginning of the training, where the density of the scene still varies significantly during optimization and where the core geometrical structure of the scene is being learned.
%As the optimization is close to convergence, the discrepancy in image quality between the pseudo-views and the original images increases, with the former becoming detrimental to the overall reconstruction, so we remove the synthesized ones from the sampling pool.

\textbf{Third and later stages}
The proposed method uses a base model $\mathbf{B}$ to augment the training data for a later model $\mathbf{D}$. This introduces a general paradigm that can be repeated iteratively until the gain saturates. Therefore, new augmented views can be synthesized with $\mathbf{D}$ and masked to train a new model $\mathbf{D}_2$, and so on. As our synthesized data augmentation regularizes the later models $\mathbf{D}_n$, these are less prone to overfitting on the initial views, which would allow us to increase the model size and reach higher quality. However, this is out of the scope of this work, as we use identical architectures across the iterations.

%% file: sec/4_experiments.tex
\section{Experiments and Results}
\label{sec:experiments}
Since having a more diverse set of views during the start of optimization benefits reconstruction, we observe faster convergence on the test views when using \pname. 
Interestingly, not only does using \pname\ improve convergence speed and test-view quality but it can also boost PSNR values on the training views, further reinforcing our intuition.

\subsection{Experimental Setup}

\textbf{Dataset and metrics}
We test our method on the challenging 7 public scenes introduced with \textbf{mip-NeRF 360}~\cite{barron2022mip360}, comprising both bounded indoor and large unbounded outdoor settings.
Additionally, we experimented with the 8 scenes of the \textbf{LLFF} dataset~\cite{mildenhall2019llff}, which includes smaller, bounded scenes. 
The former features a large amount of real images captured all around an object or scene, while the latter has fewer views of forward-facing scenes. So, the two assess complementary aspects. 
For sparsifying the mip-NeRF 360 dataset, we select the training views by sampling from all available images (including the test ones) while preserving as much coverage of the scene as possible. We then select the test views among those test views from the original set that were not selected as training images. Toward this end, we selected 30 training images (50 for \textit{stump} and \textit{bicycle} due to convergence issues of the baselines when using 30 views). The exact data splits will be made available.
With LLFF, we experimented with the highly sparse setting with just 3 views available, following the evaluation paradigm of RegNeRF~\cite{niemeyer2022regnerf}. 
We evaluate the test views on the standard metrics and errors, i.e., PSNR, SSIM~\cite{wang2004ssim}, and LPIPS~\cite{zhang2018lpips}.

%\textbf{Metrics} On top of evaluating the novel views on the standard metrics and errors, i.e., PSNR, SSIM~\cite{wang2004ssim}, and LPIPS~\cite{zhang2018lpips}, \todo{we consider the number of times each test region has been seen during training. This visibility occurrence is computed considering the overlaps of each test region in 3D space with respect to the training camera-frustums. Since accurate scene geometry is not available, we simplify the computation as follows: all patches project at a constant distance, each region is considered seen from a training frustum if its center pixel projected in 3D space is inside such frustum, and patch rays with a rotational distance higher than 30° from a frustum are not treated as visible. This last measure is to avoid that views facing one another captured from opposite sides of the scene score complete overlap. Furthermore, each image is divided into 64 equal patches, on which the PSNR is computed. While additional details on this computation are described in the supplementary material, the computation aims to report the PSNR in greater detail according to the visibility of each region.}

\input{tables/mip_table.tex}

\textbf{Prior Works and Baselines}
We showcase the effectiveness of our method by applying it on diverse NVS methods, namely PyNeRF~\cite{turki2024pynerf}, Instant-NGP~\cite{muller2022instant}, RegNeRF~\cite{niemeyer2022regnerf}, and 3DGS~\cite{kerbl2023gaussiansplat}. 3DGS uses an explicit representation, so there is no model uncertainty to estimate. 
RegNeRF, on the other hand, is implemented in JAX~\cite{jax2018github}, for which no implementation of Bayes'Rays~\cite{goli2023bayes} exists. 
Therefore, with 3DGS and RegNeRF, we show only the impact of our augmentations without uncertainty masks.
%a state-of-the-art NeRF method, namely Zip-NeRF~\cite{barron2023zip}.
%\todo{Furthermore, we compare with other methods enforcing geometric constraints, such as DS-NeRF~\cite{deng2022dsnerf}.}
%\todo{PyNeRF, 3DGS, i-NGP}

\textbf{Implementation Details}
We train our method on an image resolution down-sampled by eight from the original image sizes for the available outdoor scenes and down-sampled by four for the indoor scenes of the mip-NeRF360 dataset, bringing both to a comparable image size. For the LLFF dataset we down-sample all images by eight, following RegNeRF~\cite{niemeyer2022regnerf}. 
All models are trained using Nerfstudio~\cite{tancik2023nerfstudio}, except for RegNeRF where we use the original implementation in JAX~\cite{jax2018github}. For PyNeRF ~\cite{turki2024pynerf}, Gaussian Splatting~\cite{kerbl2023gaussiansplat} and Instant-NGP~\cite{muller2022instant}, we used their implementation available in Nerfstudio. 
We trained all Nerfstudio models for 30k iterations from scratch with a batch size of 8192 rays for PyNerf and 1024 rays for Instant-NGP and inherited all other losses and hyperparameters of the baseline methods. For RegNeRF we followed their training paradigm and hyperparameters. The influence of the synthesized views on the optimization is removed after 8k iterations for PyNerf and Gaussian-Splatting, after 5K iterations for RegNeRF and after 200 iterations for Instant-NGP.
%We inherit the distortion, anti-interlevel, and hash-decay losses that were introduced in Zip-NeRF\cite{barron2023zip} and use the publicly available unofficial PyTorch implementation\footnote{https://github.com/SuLvXiangXin/zipnerf-pytorch}, since the official one was not public at the time of this writing.}
%The $\sigma$ for our filtering is annealed logarithmically from 0.05 to 0.025 during the first 2k iterations, while our occlusion threshold is linearly decreased from 1.0 to 0.33 over the course of 20k iterations. 
We train all models on a 24GB NVIDIA RTX 3090 or 4090 GPU.

\subsection{Quantitative Results}\label{sec:quantitatives}
%In Table~\ref{tab:mip}, we report the results on the mip-NeRF 360 dataset~\cite{barron2022mip360} across a decreasing number of available training views. Specifically, we used all training inputs available (all views), half of them (1/2 views), a quarter (1/4), or an eight (1/8). We compare with Zip-NeRF~\cite{barron2023zip} as the baseline and DS-NeRF~\cite{deng2022dsnerf} applied on Zip-NeRF. The table also includes Zip-NeRF trained with 2 extra views synthesized for each available training view. Such views are generated using the baseline Zip-NeRF model (first row of the table) for each respective sparsity setting. The last row reports the results of the proposed \pname\, which, on top of training with the extra synthesized views, enforces the geometric constraints described in Section~\ref{sec:geometric_constraints}.

In Table~\ref{tab:mip}, we report the results on the mip-NeRF 360 dataset~\cite{barron2022mip360} applying the proposed \pname\ on Instant-NGP~\cite{muller2022instant}, PyNeRF~\cite{turki2024pynerf}, and 3D Gaussian Splatting~\cite{kerbl2023gaussiansplat} in sparse view settings.
In Table~\ref{tab:3_views}, we report the results on the LLFF dataset~\cite{mildenhall2019llff} applying \pname\ on RegNeRF~\cite{niemeyer2022regnerf} in highly-sparse 3 view settings.
%In Table~\ref{tab:gsplat}, we show the results of our method applied on 3D Gaussian Splatting~\cite{kerbl2023gaussiansplat}.

\input{tables/regnerf.tex}

\textbf{\pname\ as flexible add-on} The proposed method brings notable improvements across various settings, increasing PSNR by 0.86 dB over PyNeRF, 0.81 dB over 3DGS, 0.64 dB over Instant-NGP, and 0.27 dB over RegNeRF already with a single iteration, while improving SSIM and LPIPS as well. 
%In sparse settings, the relatively low number of views makes it difficult to disambiguate the scene geometry and properly render novel views; thus, increasing the number of images is particularly beneficial. \pname\ generates such views with the baseline, thereby relying on the exact same original views. Yet, our iterative augmentation strategy brings valuable benefits, providing additional viewpoints on which the methods learn better representations of the scenes. \pname\ aids in structuring the scene and reducing ambiguities.
Despite their synthesized nature, adding more training views mitigates overfitting and acts as a regularizer.
In Table~\ref{tab:3_views}, in particular, we combine \pname\ with the regularization of RegNeRF, showing how \pname\ can adapt to any NVS method. This is because it is an add-on and not an alternative to them. 
% Such results are aligned with those of other iterative frameworks from other domains, where later models outperform earlier ones thanks to an implicit regularization effect occurring in the later stages~\cite{gasperini2023md4all,chen2020naivestudent}.
%In denser settings, the number of original views is already relatively high, reducing the benefits of \pname\ (see supplementary material).
%, with good coverage of the area of interest, allowing each method to reconstruct the scene significantly better. Therefore, in such setups, the synthesized views are rather close to the original ones, so the viewpoint advantage and the information gain are small. Nevertheless, \pname\ slightly improves over each of its baselines, demonstrating its efficacy.
Overall, the efficacy of our method on both implicit (NeRF-based) and explicit (3DGS) scene representations is particularly notable as it is achieved without using extra data, additional models, or supervision, but only with already available information.

%\textbf{Combination with prior works} As shown in the tables, \pname\ is general and can be applied over any NVS method. 

\textbf{\pname\ as an iterative method}
In Tables~\ref{tab:3_views} and~\ref{tab:more_rounds}, we show the impact of our method when applying it iteratively. As described in Section~\ref{sec:second_round}, \pname\ uses renderings from a first model $\mathbf{B}$ to augment the training data of a second model $\mathbf{D}$. To further improve the output quality, the process can be repeated iteratively, by using $\mathbf{D}$ to render the augmented views. We show this in the tables, where, at each iteration, the quality improves across the board for the various pipelines and settings. This confirms that NVS helps the task of NVS. While the improvement varies from scene to scene, we notice that the performance saturates beyond 3-5 rounds. Overall, the proposed \pname\ leads to a substantial boost of \textbf{+3.45 PSNR over 3DGS}~\cite{kerbl2023gaussiansplat}, without using any extra data or external models, other than what is already available to the standard 3DGS.

\input{tables/more_rounds.tex}
\input{tables/view_selection.tex}

% aspects: MVS, early in the training where the geometry is being optimized for, resetting the training, sub-optimal views still help, naive-student
\textbf{Why it works}
As described in Section~\ref{sec:first_round}, NeRF's and 3DGS' optimization is similar to SfM approaches, as matching pixel features are triangulated in 3D. Especially in sparse settings, this is highly under-constrained with many degenerate solutions. This can lead to local minima, which fit well with the training views but not the test views due to the inconsistent geometry. 
%This can be seen very well in the 3-view setting, where without some constraints the 3D points are all on the image plane. Maybe add this, maybe not. 
With \pname, we build upon this (potentially sub-optimal) initial optimization and render synthetic views as extra constraints for later rounds. It is widely known that more viewpoints help NVS. This same concept drives \pname, too, albeit with synthesized views. Thanks to these extra constraints, some degenerate solutions are removed and the optimization converges to a better 3D structure. This can be seen in the supplementary video, with smoother rendered depth maps.
Through an ablation study on 3DGS~\cite{kerbl2023gaussiansplat}, Table~\ref{tab:why} shows the importance of: early optimization stages for the scene geometry and the impact of the pseudo views (syn.views) on that, removing the pseudo views as the scene geometry converges to not learn wrong details from them, and resetting the training at each of the \pname\ rounds. Specifically, adding the synthetic views without resetting the optimization (3rd row) does not bring any benefits and worsens LPIPS. Resetting and keeping the rendered views until the end (4th row) improves, but removing them leads to better details (last row, ours). On the contrary, adding the views after the scene geometry has already been captured (5th row) leads to mixed results compared to the baselines.
While \pname\ increases the training iterations, e.g., by 2x in the table, simply doubling the iterations performs worse (2nd row).
All together, \pname\ leads to a better global solution without adding extra info over the baseline.
%Intuitively, we can explain why \pname\ works from two perspectives: the number of views and the optimization.  Therefore, the benefits are not as good as adding more real views. This lower quality requires the removal of the pseudo-views as the model captures the scene geometry, which is coarser than the fine details, being learned at later stages.
%Confusion may have arisen as we do not add extra info over the baseline, yet achieve superior results.
This is similar to other iterative frameworks from other domains~\cite{gasperini2023md4all,chen2020naivestudent}, where resetting the optimization at each round leads to better solutions.
%We referred to data augmentation as we follow its definition of enhancing the training data diversity without collecting new data.
%In fact, we use no extra info, data, or models, that is similar to data augmentation. Beyond the added views, the reason for the improvement is linked to the reach of a better solution during the optimization.
%We mentioned data augmentation, because we enhance the training set building on the available data. Perhaps, it would be better to talk about \textit{pseudo data augmentation}.

%\textbf{\pname\ vs. more training iterations} Throughout this work, we optimized all methods for 30k iterations. \pname\ increases the total iterations, e.g., by 2x considering a single round of \pname, albeit resetting the optimization every time. In Table~\ref{tab:why}, we show that doubling the iterations for the baseline, i.e., training it for as much as 1 round of \pname\ (last row), improves. However, applying ours for 1 round is even better, showcasing the benefits of of resetting the optimization and adding the novel views.

\input{tables/why.tex}

\begin{figure*}[t]
\begin{center}
\includegraphics[width=1.0\textwidth]{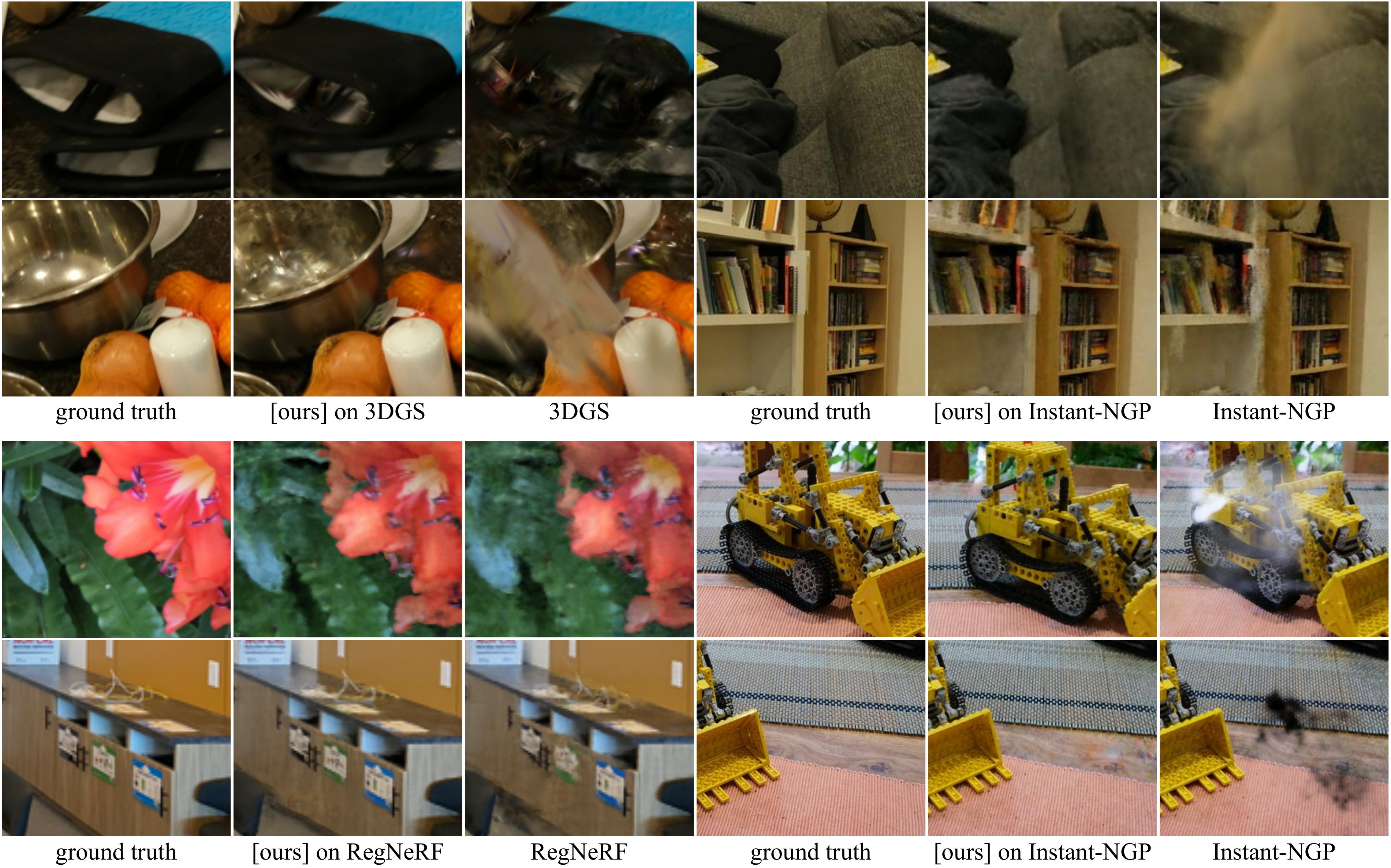}
\vspace{-0.7cm}
\end{center}
   \caption{Qualitative results on cropped images from the test set of the mip-NeRF 360 and LLFF datasets. 3DGS~\cite{kerbl2023gaussiansplat} and Instant-NGP~\cite{muller2022instant} were trained on the mip-NeRF 360 dataset in a 30 view setting, while RegNeRF~\cite{niemeyer2022regnerf} was trained on LLFF with only 3 views. Qualitative results of PyNeRF~\cite{turki2024pynerf} can be found in the supplementary material.}
\label{fig:qualitative}
\vspace{-0.25cm}
\end{figure*}
%\end{wrapfigure}

\textbf{Trade-offs with pseudo-views poses}
In Table~\ref{tab:view_selection}, we evaluate different poses for the novel view selection on PyNeRF~\cite{turki2024pynerf}, exploring different interpolations across pairs of the original views. Remarkably, all augmentations are beneficial, improving PSNR by 0.9-1.3 dB. The results show the trade-offs further described in the supplementary material: the furthest from the original views (e.g., .33-.66 is further than 0.20-0.80), the lower the quality of the novel views, and the smaller the performance improvement.
However, the opposite trade-off occurs, too: the closer to the initial images, the less amount of new information to gain for the multi-view optimization.
Therefore, it is better to sample more pseudo-views closer to the original ones while still including further viewpoints for extra information, as for 7x.

\textbf{Impact of the augmentation factor}
In Table~\ref{tab:view_selection}, we also explore different augmentation factors $N$. Higher factors are more beneficial despite the introduction of more sub-optimal, synthesized views. Doubling the views (2x) brings the largest relative gain with +0.9 dB PSNR. The performance further improves as the factor is increased. However, at high $N$ factors (e.g., 11x), benefits reduce as the influence of the many synthetic views overcomes the few original images, with 7x striking a good balance.

\textbf{Optimizing for specific views}
In Table~\ref{tab:view_selection}, we show how we can optimize for specific views, by augmenting with 4 test views synthesized by the baseline. We compare this with augmenting with 4 randomly interpolated views and evaluate on the viewpoint of the 4 test views. While adding random views helps, specifically synthesizing the viewpoints of interest (e.g., the test views) shows higher margins on all metrics. In practice, \pname\ allows to improve on viewpoints of interest, which can be beneficial when the available views lead to a sub-optimal quality in specific areas. In such settings, \pname\ can densify the views around such areas and ultimately improve the outputs.

%\textbf{Impact of the pseudo-views position}

%\todo{\textbf{Visibility-dependent evaluation}}
%In Table~\ref{tab:overlap}, we report the PSNR results of Zip-NeRF~\cite{barron2023zip} and the proposed \pname\ applied on Zip-NeRF. It shows how the proposed \pname\ performs better than Zip-NeRF regardless of how many times each region of the test images is seen during training. However, it is interesting how the gap varies. In the all-views setting, our method achieves the highest gain on those image regions that are approximately seen only once or twice during training. At higher frequencies, the gap reduces until it becomes negligible, with patches seen more than 20 times. When using only half of the training views (1/2), the margin is higher on the patches seen more often, although these are relatively few as only the \textit{room} scene exhibits regions visible more than 15 times. Overall, the table indicates how \pname\ mainly contributes to improving the novel view synthesis of the low-visibility regions. However, when fewer views are available (1/2), the gain in such rarely seen regions remains similar, while the overall scene benefits more from the multiple pseudo-views added by \pname.

\subsection{Qualitative Results}

%\textbf{Sparser settings}

Figure~\ref{fig:qualitative} shows examples of synthesized images with 3DGS~\cite{kerbl2023gaussiansplat}, RegNeRF~\cite{niemeyer2022regnerf}, and Instant-NGP~\cite{muller2022instant}, with and without our method. As already seen in the quantitative results (Section~\ref{sec:quantitatives}), the output quality of the proposed \pname\ improves significantly upon the baselines, especially in terms of the reduction of floating artifacts.
In particular, the addition of our method sharpens the scene and better preserves the small details as can be seen for RegNeRF in Figure \ref{fig:qualitative}.
%, such as the table cloth in the \textit{bonsai} scene, or the books and picture frame in the \textit{room} scene. 
Remarkably, the proposed method is able to remove dominant artifacts in the renderings by Instant-NGP and 3DGS, despite using the same original views, architecture, and hyperparameters as the baselines.
The augmented views help disambiguating the scene geometry, allowing all models to deliver better results by converging to a better solution.

The \textbf{supplementary material} includes additional details and results, such as the performance on each scene and extra qualitative examples in a video. There we also show how the proposed \pname\ can densify the SfM point cloud that can be used to initialize 3DGS. 

\textbf{Limitations}
Our method relies on the first stage to provide reasonable renderings and a decent estimate of the scene geometry, so it is challenging to enhance poorly rendered scenes, e.g., the \textit{stump} scene with Instant-NGP~\cite{muller2022instant}, which could not converge.
The proposed method also relies on the uncertainty estimates to discard artifacts and low-quality areas. Therefore, better uncertainty estimates would be beneficial. Uncertainty calibration  of the threshold $\mu$ could help refine the estimates while preserving the signal in higher-quality areas.

%% file: tables/mip_table.tex
\begin{table}[t]
\setlength{\tabcolsep}{7.8pt}
\begin{center}
\begin{tabular}{l|ccc}
\toprule
%& \multicolumn{3}{c}{sparse views} \\
Method & PSNR & SSIM & LPIPS \\
\midrule

Instant-NGP~\cite{muller2022instant} & 20.76 & 0.659 & 0.301 \\
%+ \pname\ [ours]* & \textbf{21.40} & \textbf{0.682} & \textbf{0.270} \\
+ \pname\ [ours] & \textbf{21.35}	& \textbf{0.680} & \textbf{0.273} \\

\midrule

PyNeRF~\cite{turki2024pynerf} & 22.65 & 0.746 & 0.182 \\
+ \pname\ [ours] & \textbf{23.51} & \textbf{0.772} & \textbf{0.160} \\

\midrule

3DGS~\cite{kerbl2023gaussiansplat} & 20.49 & 0.600 & 0.259 \\
+ \pname\ [ours] & \textbf{21.30} & \textbf{0.627} & \textbf{0.235} \\

\bottomrule
\end{tabular}
\vspace{-0.7em}
\caption{Rendering performance on the mip-NeRF 360~\cite{barron2022mip360} dataset in \textbf{sparse view settings} averaged over the scenes. The proposed \pname\ is applied on the Instant-NGP~\cite{muller2022instant}, PyNeRF~\cite{turki2024pynerf}, and 3DGS~\cite{kerbl2023gaussiansplat} baselines, using the corresponding baseline to augment the views. Uncertainty estimates are unavailable for 3DGS.}
\label{tab:mip}
\vspace{-1.5em}
\end{center}
\end{table}

\begin{comment}
\begin{tabular}{l|ccc|ccc}
\toprule
& \multicolumn{3}{c|}{sparser views} & \multicolumn{3}{c}{denser views} \\
Method & PSNR & SSIM & LPIPS & PSNR & SSIM & LPIPS \\
\midrule

Instant-NGP~\cite{muller2022instant} & 20.76 & 0.659 & 0.301 & 25.87 & 0.812 & 0.162 \\
%+ \pname\ [ours]* & \textbf{21.40} & \textbf{0.682} & \textbf{0.270} & \textbf{25.88} & \textbf{0.814} & \textbf{0.161} \\
+ \pname\ [ours] & \textbf{21.35}	& \textbf{0.680} & \textbf{0.273} \\

\midrule

PyNeRF~\cite{turki2024pynerf} & 22.65 & 0.746 & 0.182 & 27.41 & 0.874 & 0.094\\
+ \pname\ [ours] & \textbf{23.51} & \textbf{0.772} & \textbf{0.160} & \textbf{27.43} & \textbf{0.875} & \textbf{0.093}\\

\midrule

3DGS~\cite{kerbl2023gaussiansplat} & 20.49 & 0.600 & 0.259 & 27.83 & 0.855 & \textbf{0.109} \\
+ \pname\ [ours]* & \textbf{21.30} & \textbf{0.627} & \textbf{0.235} & \textbf{28.03} & \textbf{0.859} & 0.111 \\

\bottomrule
\end{tabular}
\end{comment}

%% file: tables/regnerf.tex
\begin{table}[t]
\setlength{\tabcolsep}{5.9pt}
\begin{center}
\begin{tabular}{l|ccc}
\toprule
\multicolumn{1}{l|}{\textbf{3 Views}} & PSNR & SSIM & LPIPS \\
\midrule

baseline: mip-NeRF~\cite{barron2021mip} & 14.62 & 0.351 & 0.495 \\
+ RegNeRF~\cite{niemeyer2022regnerf} & 19.08 & 0.685 &  0.148\\
+ \pname, 1 round [ours] & 19.35 & 0.713 &  0.133\\
+ extra round, 2nd [ours] & \uline{19.47} & \uline{0.724} &  \textbf{0.126}\\
+ extra round, 3rd [ours] & \textbf{19.52} & \textbf{0.725} & \uline{0.127} \\

\bottomrule
\end{tabular}
\vspace{-0.7em}
\caption{Highly sparse settings with only \textbf{3 views} on the LLFF dataset~\cite{mildenhall2019llff}. We apply ours on top of RegNeRF~\cite{niemeyer2022regnerf}. Then, we iteratively apply ours via the extra rounds. The results are averaged over all LLFF scenes.}
\label{tab:3_views}
\vspace{-1.5em}
\end{center}
\end{table}

%% file: tables/more_rounds.tex
\begin{table}[t]
\setlength{\tabcolsep}{3.1pt}
\begin{center}
\begin{tabular}{l|ccc|ccc}
\toprule
& \multicolumn{3}{c|}{3DGS~\cite{kerbl2023gaussiansplat}} & \multicolumn{3}{c}{PyNeRF~\cite{turki2024pynerf}} \\
%Method & PSNR $\uparrow$ & SSIM $\uparrow$ & LPIPS $\downarrow$ \\
Method & PSNR & SSIM & LPIPS & PSNR & SSIM & LPIPS \\
\midrule

baseline & 19.31 & 0.609 & 0.296 & 17.57 & 0.643 & 0.325 \\
%baseline, 2x iterations & 19.68 & 0.633 & 0.274 & 17.93	& 0.653 & 0.331 \\
+ [ours]: 1 & 21.08 & 0.686 & 0.241 & 18.09 & 0.676 & 0.299 \\
+ [ours]: 2 & 21.89 & 0.720 & 0.214 & 18.36 & 0.693 & 0.285 \\
+ [ours]: 3 & 22.47 & 0.748 & 0.191 & 18.41 & 0.699 & 0.281 \\
+ [ours]: 4 & \uline{22.63} & \uline{0.756} & \uline{0.184} & \uline{18.47} & \uline{0.704} & \uline{0.275} \\
+ [ours]: 5 & \textbf{22.76} & \textbf{0.761} & \textbf{0.178} & \textbf{18.64} & \textbf{0.708} & \textbf{0.272} \\

\bottomrule
\end{tabular}
\vspace{-0.7em}
\caption{Rendering performance on the mip-NeRF 360~\cite{barron2022mip360} \textit{counter} scene in the 30-view setting. \pname\ is applied iteratively across multiple rounds, from 1 to 5.
%and the effect of training the baseline for double the amount of iterations is also shown (2x). All other models in this work were trained for 30k iterations.
}
\label{tab:more_rounds}
\vspace{-1.5em}
\end{center}
\end{table}

%% file: tables/view_selection.tex
%\begin{wraptable}{l}{1.0\linewidth}
\begin{table}[b]
%\vspace{-0.2em}
%\vspace{-0.35cm}
%\setlength{\tabcolsep}{5pt}
\begin{center}
\begin{tabular}{l|ccc}
\toprule
%& \multicolumn{3}{c|}{dense views} & \multicolumn{3}{c}{sparse views} \\
%Method & PSNR $\uparrow$ & SSIM $\uparrow$ & LPIPS $\downarrow$ \\
Method & PSNR & SSIM & LPIPS \\
\midrule

1x: PyNeRF~\cite{turki2024pynerf}, baseline & 22.42 & 0.632 & 0.218 \\
\midrule
1.08x: 4 random syn.views & 22.58 & 0.653 & 0.205 \\
1.08x: 4 test syn.views & \textbf{22.64} & \textbf{0.663} & \textbf{0.196} \\
\midrule
%Zip-NeRF + DS-NeRF~\cite{deng2022dsnerf} & 30.16 & 0.903 & 0.058 & - & - & - & - & - & - & - & - & - \\
%\midrule
2x: .50 & 23.35 & 0.669 & 0.190 \\
3x: .33-.66 & 23.40 & 0.666 & 0.191 \\
3x: .20-.80 & 23.61 & 0.670 & 0.189\\
3x: .10-.90 & 23.47 & 0.670 & 0.191 \\
4x: .25-.50-.75 & 23.51 & 0.667 & 0.192 \\
5x: .2-.4-.6-.8 & 23.63 & 0.668 & 0.189 \\
7x: 5x + .1-.9 [selected] & \textbf{23.73} & \textbf{0.672} & \textbf{0.184} \\
11x: 7x + .05-.3-.5-.7-.95 & 23.48 & 0.669 & 0.189 \\ %5-10-20-30-40-50-60-70-80-90-95
%\midrule
%7x + masks & 23.76 & 0.672 & 0.188 \\

\bottomrule
\end{tabular}
\vspace{-0.2em}
\caption{View selection strategies on the \textit{stump} 50-view setting of the mip-NeRF 360~\cite{barron2022mip360} dataset. Models based on PyNeRF~\cite{turki2024pynerf}. Our augmentation factors $N$ from 2 to 11 are evaluated.}
%11x corresponds to the $\alpha$ factors .05-.1-.2-.3-.4-.5-.6-.7-.8-.9-.95.
\vspace{-1.1em}
\label{tab:view_selection}
\end{center}
\end{table}
%\end{wraptable}

%% file: tables/why.tex
\begin{table}[t]
\setlength{\tabcolsep}{5.2pt}
%\vspace{-1.3em}
\begin{center}
\begin{tabular}{l|ccc}
\toprule
\multicolumn{1}{l|}{sparse views} & PSNR & SSIM & LPIPS \\
\midrule

baseline: 3DGS~\cite{kerbl2023gaussiansplat}, 30k iter. & 19.31 & 0.609 & 0.296 \\
2x longer training (60k iter.) & 19.68 & 0.633 & 0.274 \\
no reset, add syn.views at 30k & 19.67 & 0.631 & 0.333\\
reset, add and keep syn.views & 20.15 & 0.654 & 0.291\\
reset, add syn.views at 15k & 19.79 & 0.637 & 0.325\\
\uline{reset, remove syn.views at 8k} & \textbf{21.08} & \textbf{0.686} & \textbf{0.241} \\

\bottomrule
\end{tabular}
\vspace{-0.7em}
\caption{Ablation study. 30-view setting, \textit{counter} scene of the mip-NeRF 360 dataset~\cite{barron2022mip360} with 3DGS~\cite{kerbl2023gaussiansplat}. The last row represents the proposed method.}
\label{tab:why}
\vspace{-1.5em}
\end{center}
\end{table}

%% file: sec/5_conclusion.tex
\section{Conclusion}
\label{sec:conclusion}
This work enhances NeRFs and 3DGSs by leveraging their own novel view synthesis capabilities.
The introduced \pname\ is a general approach that synthesizes novel views to improve the scene coverage and augment the training data for a new model. To provide a valuable training signal for the second model, \pname\ computes the NeRF's uncertainty for each augmented view to discard the uncertain regions and reduce artifacts.
As shown with various pipelines, our method yields improvements in both reconstruction and rendering quality in sparse scenes, mitigating the need for extensive captures of the scene.

%% file: sec/X_suppl.tex
\clearpage
\setcounter{page}{1}
\appendix
\maketitlesupplementary
\section{Appendix}

In this supplementary material, we complement the main paper with additional details and results.
%Specifically, in Section~\ref{sec:appendix_more_rounds}, we show the iterative power of the proposed method, in Section~\ref{sec:appendix_vs_longer_training}, we show the difference between adding our augmented views and training again instead of training for longer, in 
Specifically, we illustrate and discuss the trade-offs due to the synthetic view poses (Section~\ref{sec:suppl_trade-offs}), we experiment with and discuss the impact of our method on the SfM point cloud (Section~\ref{sec:suppl_sfm}), we present the result in dense-view settings (Section~\ref{sec:suppl_dense}), we discuss an extra ablation study (Section~\ref{sec:suppl_ablation}), we show more qualitative results, also through a video (Section~\ref{sec:suppl_qualitatives}), and we present the quantitative results by scene (Section~\ref{sec:suppl_by_scene}).

%\subsection{Optimizing for Specific Views}\label{sec:appendix_specific_views}

%In Table~\ref{tab:more_views}, we show how our method offers the opportunity of optimizing for specific views. Toward this end, we train PyNeRF~\cite{turki2024pynerf} augmenting with 4 randomly interpolated views synthesized by the baseline and compare it with another PyNeRF model trained augmenting with 4 test views synthesized by the same baseline. Both models are evaluated on the viewpoint of the 4 test views. While adding the randomly interpolated views benefits the output quality, specifically synthesizing the viewpoints of interest (i.e., the test views in this case) shows a higher margin on all metrics. In practice, this means that \pname\ allows to improve on specific viewpoints of interest, which can be beneficial when the views available lead to a sub-optimal quality from specific areas. In such setting, \pname\ can be used to densify the viewpoints around the area lacking in terms of quality, and ultimately improve the novel view synthesis.

\input{tables/dense.tex}

\subsection{More on the Views Positioning Trade-Offs}
\label{sec:suppl_trade-offs}
\begin{figure*}[t]
\begin{center}
\includegraphics[width=1.00\textwidth]{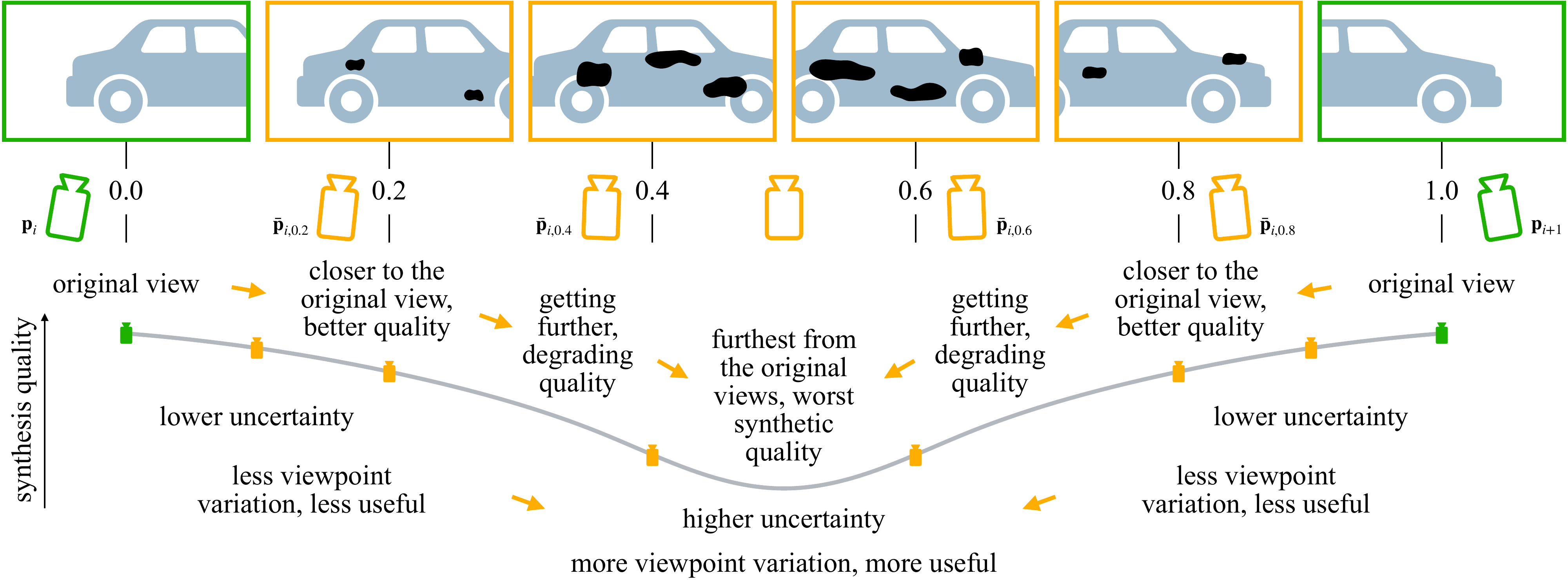}
%\vspace{-0.7cm}
\end{center}
   \caption{Illustration of the trade-offs arising when synthesizing novel views (orange) in between existing ones (green). The numbers indicate the interpolation factors. The closer to the original views, the higher the quality of the synthesis and the lower the uncertainty. Moving further from the original views (i.e., toward 0.5), the quality degrades and the uncertainty increases. Yet, an opposite trade-off occurs as views too close to the original ones do not bring any extra information. We balance this by using many views and removing artifacts thanks to the uncertainty estimates.}
\label{fig:trade_offs_distance_quality}
%\vspace{-0.2cm}
\end{figure*}

In Figure~\ref{fig:trade_offs_distance_quality}, we illustrate the trade-offs associated with the relative positioning between the original views and the synthesized ones. Considering two existing views, the quality of the views synthesized between them varies as follows: the closer the synthetic views are to the original views, the higher their quality because the viewpoint change is small. Instead, the further from the original views (e.g., towards 0.5), the worse the quality becomes (e.g., more artifacts) as the viewpoint shift is more significant and the uncertainty increases. However, synthesizing views only in the vicinity of the original ones is not particularly beneficial as the views are too similar to the original ones and, as such, do not help disambiguate the scene geometry. Therefore, the views should also be synthesized at further locations. With \pname, to mitigate the quality degradation, we remove the artifacts within the synthesized views (i.e., occurring more at the further poses) by filtering out the uncertain regions. This allows us to synthesize many views throughout the interpolated poses. We sample more poses closer to the original views (i.e., 0.0 and 1.0) and fewer poses in the middle. Specifically, we synthesize in the 7x configuration at the interpolation factors 0.1, 0.2, 0.4, 0.6, 0.8, and 0.9.
This trade-off is explored in Table 4. %~\ref{tab:view_selection}

\begin{figure}[t]
\begin{center}
\includegraphics[width=1.0\linewidth]{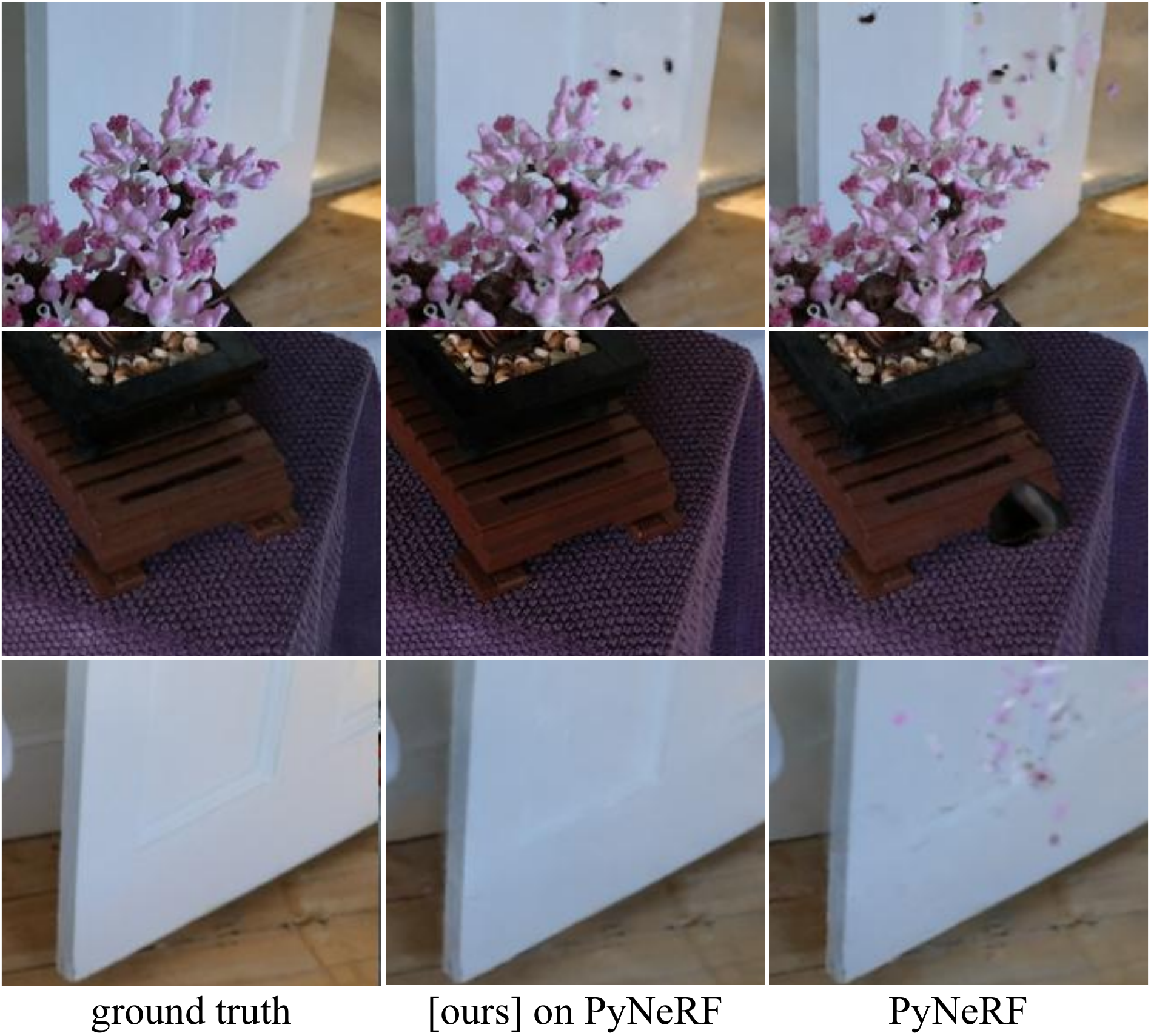}
\vspace{-0.7cm}
\end{center}
   \caption{Qualitative results on cropped images from the test set of the mip-NeRF 360 dataset~\cite{barron2022mip360} showcasing PyNeRF~\cite{turki2024pynerf} in sparse-view settings, with and without the proposed \pname.}
\label{fig:qualitatives_pynerf}
\vspace{-0.25cm}
\end{figure}

\input{tables/sfm_gs.tex}

\subsection{Impact on the SfM Point Cloud}
\label{sec:suppl_sfm}
In Table~\ref{tab:sfm_gs}, we investigate the impact of \pname\ on the SfM point cloud. For these experiments, we used 3DGS~\cite{kerbl2023gaussiansplat} initialized with the SfM point cloud constructed by the SfM pipeline of Nerfstudio~\cite{tancik2023nerfstudio} using Hloc~\cite{sarlin2019coarse} and Superpoint~\cite{detone2018superpoint} from the sparse settings. For the second stage of \pname, we followed the same protocol of Section 3 but ran the aforementioned SfM-pipeline on the synthetic and real images combined.
%\ref{sec:method}
Remarkably, despite synthetic views, not only did \pname\ improve over the standard 3DGS but also densified the point cloud by a substantial 4.6x on average. This significant difference in the number of points (\# points) is evidence of the effectiveness of NVS and \pname\ to aid and simplify the triangulation of the scene geometry, which is the foundation of our work. In particular, the \textit{bicycle} scene has more points than the other ones because of its higher image resolution.
Thanks to a much denser point cloud as initialization and the added synthesized views, the proposed \pname\ reaches a better solution, as shown by the metrics.

\subsection{Dense-View Settings}
\label{sec:suppl_dense}
Throughout this work, we experimented with challenging sparse-view settings. There, the limited views available prevent a good triangulation of the scene geometry. In Table~\ref{tab:mip_dense}, we show the impact of our \pname\ on Instant-NGP~\cite{muller2022instant}, PyNeRF~\cite{turki2024pynerf}, and 3DGS~\cite{kerbl2023gaussiansplat} in dense-view settings. Therefore, in these experiments, we use all the available training views. Remarkably, despite the already many views available, by adding synthesized ones, the proposed \pname\ does not degrade the performance and even slightly improves it. Nearly equal results are seen for Instant-NGP and PyNeRF, while a +0.2db PSNR is brought over 3DGS. This shows the flexibility of \pname.

\input{tables/ablation.tex}

\subsection{Additional Ablation Study}
\label{sec:suppl_ablation}
In Table~\ref{tab:ablation}, we report an ablation study for both PyNeRF~\cite{turki2024pynerf} and Instant-NGP~\cite{muller2022instant} in the 30-view setting. Training again with the synthetic views produced by the baseline (with $N=7$) slightly degrades the performance for all metrics for PyNeRF, while it improves for Instant-NGP. The degradation for PyNeRF is due to the lower quality signal of 6x more views than the reliable signal of the original views. Specifically, the model learns the artifacts of the augmented views and underperforms. Instead, for Instant-NGP, the quality of the augmented views was higher, leading to improvements.
Nevertheless, removing the images from the sampling pool once the scene structure has been mainly learned (stop, 3rd row) prevents both models from overfitting to the artifacts and leads to better results on all metrics, with large improvements for PyNeRF. Furthermore, masking out the uncertain regions of such synthetic views improves even further across the board for both methods.
This shows once again how synthetic views of the same scene are beneficial aid the novel view synthesis task. In particular, the results confirm the importance of increasing the scene coverage and that of having a good training signal, testifying the need for the uncertainty-based masks and our augmentation strategy.

%\input{tables/more_views.tex}

%\subsection{3DGS in Denser Settings}\label{sec:appendix_gsplat_denser}

\subsection{Additional Qualitative Results}
\label{sec:suppl_qualitatives}
\textbf{PyNeRF} In Figure~\ref{fig:qualitatives_pynerf}, we showcase the impact of our method on PyNeRF~\cite{turki2024pynerf}. As for the other methods seen in the main paper, our \pname\ reduces artifacts and leads to a better solution, e.g., by removing floaters. Examples are the floating leaves of the bonsai (first and last rows), or the black blob (second row).

\textbf{Video} We refer to the attached supplementary video
%\url{https://youtu.be/8mo3s1XAwHQ} 
for more qualitative results. The video interpolates across the test views for models trained in sparse settings. Remarkably, when paired with the proposed \pname\, 3DGS~\cite{kerbl2023gaussiansplat} reaches significantly higher rendering quality in the \textit{counter} scene (30 views). Thanks to our method, the scene geometry is substantially better captured by the Gaussians, eliminating major structural artifacts. As shown in Table~\ref{tab:garden}, the \textit{garden} scene is easier to render. In this case, our method helps with the scene geometry, especially visible in the rendered depths, as it eliminates the holes in the ground.

\subsection{Evaluation by Scene}
\label{sec:suppl_by_scene}
\textbf{LLFF} In Tables from~\ref{tab:llff_fortress} to~\ref{tab:llff_trex}, we report the performance for each scene in the 3-view setting of LLFF~\cite{mildenhall2019llff} for RegNeRF~\cite{niemeyer2022regnerf} and \pname. The detailed results showcase that the proposed method always improves upon RegNeRF, while the performance sometimes saturates before 3 rounds.

\textbf{mip-NeRF 360} In Tables from~\ref{tab:counter} to~\ref{tab:room}, we report the performance for each scene for the various pipelines on which we applied \pname, namely Instant-NGP~\cite{muller2022instant}, PyNeRF~\cite{turki2024pynerf}, and 3D Gaussian Splatting~\cite{kerbl2023gaussiansplat}. The detailed results showcase the benefits of our method indoor and outdoor, for sparser and denser view settings, and for both implicit and explicit methods.

\input{tables/by_scene_llff/fortress}
\input{tables/by_scene_llff/horns}
\input{tables/by_scene_llff/fern}
\input{tables/by_scene_llff/flower}
\input{tables/by_scene_llff/orchids}
\input{tables/by_scene_llff/leaves}
\input{tables/by_scene_llff/room}
\input{tables/by_scene_llff/trex}

\input{tables/by_scene/counter.tex}
\input{tables/by_scene/bicycle.tex}
\input{tables/by_scene/stump.tex}
\input{tables/by_scene/garden.tex}
\input{tables/by_scene/bonsai.tex}
\input{tables/by_scene/kitchen.tex}
\input{tables/by_scene/room.tex}

%% file: tables/dense.tex
\begin{table}[b]
\setlength{\tabcolsep}{7.8pt}
\begin{center}
\begin{tabular}{l|ccc}
\toprule
%& \multicolumn{3}{c}{denser views} \\
Method & PSNR & SSIM & LPIPS \\
\midrule

Instant-NGP~\cite{muller2022instant} & 25.87 & 0.812 & 0.162 \\
%+ \pname\ [ours]* & \textbf{21.40} & \textbf{0.682} & \textbf{0.270} & \textbf{25.88} & \textbf{0.814} & \textbf{0.161} \\
+ \pname\ [ours] & \textbf{25.88} & \textbf{0.814} & \textbf{0.161} \\

\midrule

PyNeRF~\cite{turki2024pynerf} & 27.41 & 0.874 & 0.094\\
+ \pname\ [ours] & \textbf{27.43} & \textbf{0.875} & \textbf{0.093}\\

\midrule

3DGS~\cite{kerbl2023gaussiansplat} & 27.83 & 0.855 & \textbf{0.109} \\
+ \pname\ [ours] & \textbf{28.03} & \textbf{0.859} & 0.111 \\

\bottomrule
\end{tabular}
%\vspace{-0.7em}
\caption{Rendering performance on the mip-NeRF 360~\cite{barron2022mip360} dataset in \textbf{dense view settings} averaged over the scenes. The proposed \pname\ is applied on the Instant-NGP~\cite{muller2022instant}, PyNeRF~\cite{turki2024pynerf}, and 3DGS~\cite{kerbl2023gaussiansplat} baselines, using the corresponding baseline to augment the views. Uncertainty estimates are unavailable for 3DGS.}
\vspace{-1.1em}
\label{tab:mip_dense}
%\vspace{-1.5em}
\end{center}
\end{table}

%% file: tables/sfm_gs.tex
\begin{table*}[t]
\setlength{\tabcolsep}{8pt}
\begin{center}
\begin{tabular}{l|cccc|cccc}
\toprule
& \multicolumn{4}{c|}{3DGS~\cite{kerbl2023gaussiansplat} (SfM, Hloc+Superpoint)}  & \multicolumn{4}{c}{\pname\ (SfM, Hloc+Superpoint)} \\
Scene - \textit{sparse} & PSNR & SSIM & LPIPS & \# Points & PSNR & SSIM & LPIPS & \# Points\\
\midrule
\textit{bonsai} & 26.19 & 0.847 & 0.131 & 6683 & 26.52 & 0.864 & 0.126 & 20274 \\
\textit{counter} & 23.29 & 0.766 & 0.167 & 8032 & 23.32 & 0.773 & 0.165 & 30855 \\
\textit{kitchen} & 24.27 & 0.849 & 0.098 & 3914 & 24.22 & 0.852 & 0.099 & 13716\\
\textit{room} & 24.82 & 0.829 & 0.162 & 4304 & 24.73 & 0.833 & 0.167 & 16216 \\
\textit{garden} & 24.33 & 0.773 & 0.089 & 4873 & 24.87 & 0.793 & 0.081 & 13646 \\ 
\textit{stump} & 21.97 & 0.528 & 0.230 & 7337 & 22.36 & 0.568 & 0.228 & 38891 \\ 
\textit{bicycle} & 20.04 & 0.482 & 0.329 & 18831 & 20.38 & 0.514 & 0.301 & 114554 \\
\midrule
average & 23.56 & 0.725 & 0.172 & 7711 & \textbf{23.77} & \textbf{0.742} & \textbf{0.167} & \textbf{35450} \\

\bottomrule
\end{tabular}
%\vspace{-0.7em}
\caption{Impact of \pname\ on the SfM point cloud and 3DGS~\cite{kerbl2023gaussiansplat} initialized with it. SfM is run for all with HLoc~\cite{sarlin2019coarse} with Superpoint~\cite{detone2018superpoint} features. For \pname, SfM is run with the synthesized views, too.}
\label{tab:sfm_gs}
%\vspace{-1.5em}
\end{center}
\end{table*}

%% file: tables/ablation.tex
\begin{table*}[t]
\setlength{\tabcolsep}{8pt}
\begin{center}
\begin{tabular}{l|ccc|ccc}
\toprule
& \multicolumn{3}{c|}{PyNeRF~\cite{turki2024pynerf}} & \multicolumn{3}{c}{Instant-NGP~\cite{muller2022instant}} \\
Method & PSNR & SSIM & LPIPS & PSNR & SSIM & LPIPS \\
\midrule

baseline & 23.16 & 0.840 & 0.119 & 20.75 & 0.709 & 0.232 \\
+ augmented views & 23.08 & 0.829 & 0.125 & 21.54 & 0.740 & 0.208 \\
+ stop aug.~views & 23.84 & 0.862 & 0.102 & 21.57 & 0.743 & 0.201 \\
+ masks & \textbf{24.08} & \textbf{0.872} & \textbf{0.099} & \textbf{21.70} & \textbf{0.753} & \textbf{0.195} \\

\bottomrule
\end{tabular}
%\vspace{-0.7em}
\caption{Ablation on the mip-NeRF 360~\cite{barron2022mip360} dataset in the sparser settings of the \textit{kitchen} scene (30 views). The proposed \pname\ is applied on Instant-NGP~\cite{muller2022instant} and PyNeRF~\cite{turki2024pynerf}, using the corresponding baseline to augment the views.}
\label{tab:ablation}
%\vspace{-1.5em}
\end{center}
\end{table*}

%% file: tables/by_scene_llff/fortress.tex
\begin{table}[t]
\begin{center}
\begin{tabular}{l|ccc}
\toprule
\multicolumn{1}{l|}{\textit{fortress}, 3 views} & PSNR & SSIM & LPIPS \\
\midrule

RegNeRF~\cite{niemeyer2022regnerf} & 23.32 & 0.743 & 0.065\\
+ \pname, 1 round [ours] & 23.69 & 0.789 & 0.054\\
+ extra round, 2nd [ours] & 23.94 & 0.814 & 0.047\\
+ extra round, 3rd [ours] & \textbf{24.34} & \textbf{0.829} & \textbf{0.046} \\

\bottomrule
\end{tabular}
%\vspace{-0.7em}
\caption{Results on the \textit{fortress} scene of the LLFF dataset~\cite{mildenhall2019llff} for the \textbf{3-view} settings. The proposed \pname\ is applied on RegNeRF~\cite{niemeyer2022regnerf} and then to itself iteratively (rounds).}
\label{tab:llff_fortress}
%\vspace{-1.5em}
\end{center}
\end{table}

%% file: tables/by_scene_llff/horns.tex
\begin{table}[t]
\begin{center}
\begin{tabular}{l|ccc}
\toprule
\multicolumn{1}{l|}{\textit{horns}, 3 views} & PSNR & SSIM & LPIPS \\
\midrule

RegNeRF~\cite{niemeyer2022regnerf} & 15.65 & 0.610 & 0.200\\
+ \pname, 1 round [ours] & 15.75 & 0.654 & 0.179\\
+ extra round, 2nd [ours] & 15.80 & 0.655 & 0.168\\
+ extra round, 3rd [ours] & \textbf{16.68} & \textbf{0.670} & \textbf{0.161} \\

\bottomrule
\end{tabular}
%\vspace{-0.7em}
\caption{Results on the \textit{horns} scene of the LLFF dataset~\cite{mildenhall2019llff} for the \textbf{3-view} settings. The proposed \pname\ is applied on RegNeRF~\cite{niemeyer2022regnerf} and then to itself iteratively (rounds).}
\label{tab:llff_horns}
%\vspace{-1.5em}
\end{center}
\end{table}

%% file: tables/by_scene_llff/fern.tex
\begin{table}[t]
\begin{center}
\begin{tabular}{l|ccc}
\toprule
\multicolumn{1}{l|}{\textit{fern}, 3 views} & PSNR & SSIM & LPIPS \\
\midrule

RegNeRF~\cite{niemeyer2022regnerf} & 19.87 & 0.697 & 0.202\\
+ \pname, 1 round [ours] & \textbf{19.96} & 0.709 & 0.191\\
+ extra round, 2nd [ours] & \textbf{19.96} & 0.714 & \textbf{0.180}\\
+ extra round, 3rd [ours] & 19.95 & \textbf{0.718} & \textbf{0.180} \\

\bottomrule
\end{tabular}
%\vspace{-0.7em}
\caption{Results on the \textit{fern} scene of the LLFF dataset~\cite{mildenhall2019llff} for the \textbf{3-view} settings. The proposed \pname\ is applied on RegNeRF~\cite{niemeyer2022regnerf} and then to itself iteratively (rounds).}
\label{tab:llff_fern}
%\vspace{-1.5em}
\end{center}
\end{table}

%% file: tables/by_scene_llff/flower.tex
\begin{table}[t]
\begin{center}
\begin{tabular}{l|ccc}
\toprule
\multicolumn{1}{l|}{\textit{flower}, 3 views} & PSNR & SSIM & LPIPS \\
\midrule

RegNeRF~\cite{niemeyer2022regnerf} & 19.93 & 0.688 & 0.156\\
+ \pname, 1 round [ours] & 20.28 & 0.711 & 0.143\\
+ extra round, 2nd [ours] & \textbf{20.46} & \textbf{0.720} & \textbf{0.140}\\
+ extra round, 3rd [ours] & 20.04 & 0.695 & 0.154 \\

\bottomrule
\end{tabular}
%\vspace{-0.7em}
\caption{Results on the \textit{flower} scene of the LLFF dataset~\cite{mildenhall2019llff} for the \textbf{3-view} settings. The proposed \pname\ is applied on RegNeRF~\cite{niemeyer2022regnerf} and then to itself iteratively (rounds).}
\label{tab:llff_flower}
%\vspace{-1.5em}
\end{center}
\end{table}

%% file: tables/by_scene_llff/orchids.tex
\begin{table}[t]
\begin{center}
\begin{tabular}{l|ccc}
\toprule
\multicolumn{1}{l|}{\textit{orchids}, 3 views} & PSNR & SSIM & LPIPS \\
\midrule

RegNeRF~\cite{niemeyer2022regnerf} & 15.56 & 0.502 & 0.177\\
+ \pname, 1 round [ours] & 15.77 & 0.526 & 0.162\\
+ extra round, 2nd [ours] & 15.86 & 0.540 & 0.155\\
+ extra round, 3rd [ours] & \textbf{15.90} & \textbf{0.547} & \textbf{0.150} \\

\bottomrule
\end{tabular}
%\vspace{-0.7em}
\caption{Results on the \textit{orchids} scene of the LLFF dataset~\cite{mildenhall2019llff} for the \textbf{3-view} settings. The proposed \pname\ is applied on RegNeRF~\cite{niemeyer2022regnerf} and then to itself iteratively (rounds).}
\label{tab:llff_orchids}
%\vspace{-1.5em}
\end{center}
\end{table}

%% file: tables/by_scene_llff/leaves.tex
\begin{table}[t]
\begin{center}
\begin{tabular}{l|ccc}
\toprule
\multicolumn{1}{l|}{\textit{leaves}, 3 views} & PSNR & SSIM & LPIPS \\
\midrule

RegNeRF~\cite{niemeyer2022regnerf} & 16.60 & 0.612 & 0.132\\
+ \pname, 1 round [ours] & 16.92 & 0.654 & 0.116\\
+ extra round, 2nd [ours] & 17.07 & 0.670 & 0.107\\
+ extra round, 3rd [ours] & \textbf{17.14} & \textbf{0.677} & \textbf{0.103} \\

\bottomrule
\end{tabular}
%\vspace{-0.7em}
\caption{Results on the \textit{leaves} scene of the LLFF dataset~\cite{mildenhall2019llff} for the \textbf{3-view} settings. The proposed \pname\ is applied on RegNeRF~\cite{niemeyer2022regnerf} and then to itself iteratively (rounds).}
\label{tab:llff_leaves}
%\vspace{-1.5em}
\end{center}
\end{table}

%% file: tables/by_scene_llff/room.tex
\begin{table}[t]
\begin{center}
\begin{tabular}{l|ccc}
\toprule
\multicolumn{1}{l|}{\textit{room}, 3 views} & PSNR & SSIM & LPIPS \\
\midrule

RegNeRF~\cite{niemeyer2022regnerf} & 21.53 & 0.861 & 0.116\\
+ \pname, 1 round [ours] & 21.84 & 0.873 & 0.105\\
+ extra round, 2nd [ours] & \textbf{21.96} & \textbf{0.878} & \textbf{0.098}\\
+ extra round, 3rd [ours] & 21.47 & 0.871 & 0.107 \\

\bottomrule
\end{tabular}
%\vspace{-0.7em}
\caption{Results on the \textit{room} scene of the LLFF dataset~\cite{mildenhall2019llff} for the \textbf{3-view} settings. The proposed \pname\ is applied on RegNeRF~\cite{niemeyer2022regnerf} and then to itself iteratively (rounds).}
\label{tab:llff_room}
%\vspace{-1.5em}
\end{center}
\end{table}

%% file: tables/by_scene_llff/trex.tex
\begin{table}[t]
\begin{center}
\begin{tabular}{l|ccc}
\toprule
\multicolumn{1}{l|}{\textit{t-rex}, 3 views} & PSNR & SSIM & LPIPS \\
\midrule

RegNeRF~\cite{niemeyer2022regnerf} & 20.16 & 0.766 & 0.133\\
+ \pname, 1 round [ours] & 20.55 & 0.788 & 0.117\\
+ extra round, 2nd [ours] & \textbf{20.70} & \textbf{0.797} & \textbf{0.111}\\
+ extra round, 3rd [ours] & 20.63 & 0.796 & 0.113 \\

\bottomrule
\end{tabular}
%\vspace{-0.7em}
\caption{Results on the \textit{t-rex} scene of the LLFF dataset~\cite{mildenhall2019llff} for the \textbf{3-view} settings. The proposed \pname\ is applied on RegNeRF~\cite{niemeyer2022regnerf} and then to itself iteratively (rounds).}
\label{tab:llff_trex}
%\vspace{-1.5em}
\end{center}
\end{table}

%% file: tables/by_scene/counter.tex
\begin{table*}
\begin{center}
\begin{tabular}{l|ccc|ccc}
\toprule
& \multicolumn{3}{c|}{\textit{counter} - sparser} & \multicolumn{3}{c}{\textit{counter} - denser} \\
%Method & PSNR $\uparrow$ & SSIM $\uparrow$ & LPIPS $\downarrow$ \\
Method & PSNR & SSIM & LPIPS & PSNR & SSIM & LPIPS \\
\midrule

Instant-NGP~\cite{muller2022instant} & 17.72 & 0.557 & 0.458 & \textbf{25.52} & 0.801 & 0.183 \\
+ \pname\ [ours] & \textbf{17.97} & \textbf{0.573} & \textbf{0.422} & 25.50 & \textbf{0.805} & \textbf{0.180} \\
\midrule
PyNeRF~\cite{turki2024pynerf} & 17.57 & 0.643 & 0.325 & 26.80 & 0.864 & 0.111 \\
+ \pname\ [ours] & \textbf{18.09} & \textbf{0.676} & \textbf{0.299} & \textbf{27.13} & \textbf{0.871} & \textbf{0.107} \\
\midrule
3DGS~\cite{kerbl2023gaussiansplat} & 19.31 & 0.609 & 0.296 & 27.69 & \textbf{0.886} & \textbf{0.119} \\
+ \pname\ [ours] & \textbf{21.08} & \textbf{0.686} & \textbf{0.241} & \textbf{27.70} & 0.879 & 0.127 \\

\bottomrule
\end{tabular}
\vspace{0.2cm}
\caption{Results on the \textit{counter} scene of the mip-NeRF 360 dataset~\cite{barron2022mip360}, for sparser and denser settings. The proposed \pname\ is applied on Instant-NGP~\cite{muller2022instant}, PyNeRF~\cite{turki2024pynerf}, and 3DGS~\cite{kerbl2023gaussiansplat}. Uncertainty estimates are unavailable for 3DGS.}
\label{tab:counter}
%\vspace{-0.2cm}
\end{center}
\end{table*}

%% file: tables/by_scene/bicycle.tex
\begin{table*}
\begin{center}
\begin{tabular}{l|ccc|ccc}
\toprule
& \multicolumn{3}{c|}{\textit{bicycle} - sparser} & \multicolumn{3}{c}{\textit{bicycle} - denser} \\
%Method & PSNR $\uparrow$ & SSIM $\uparrow$ & LPIPS $\downarrow$ \\
Method & PSNR & SSIM & LPIPS & PSNR & SSIM & LPIPS \\
\midrule

Instant-NGP~\cite{muller2022instant} & 20.98 & 0.477 & 0.370  & 23.30 & 0.625 & 0.301  \\
+ \pname\ [ours] & \textbf{21.27} & \textbf{0.499} & \textbf{0.345}  & \textbf{23.35} & \textbf{0.630} & \textbf{0.300}  \\
\midrule
PyNeRF~\cite{turki2024pynerf} & 22.13 & 0.608 & 0.234  & 24.92 & 0.779 & 0.154  \\
+ \pname\ [ours] & \textbf{22.96} & \textbf{0.643} & \textbf{0.200}  & \textbf{25.19} & \textbf{0.784} & \textbf{0.150}  \\
\midrule
3DGS~\cite{kerbl2023gaussiansplat} & 17.48 & 0.320 & 0.403  & 25.37 & \textbf{0.775} & \textbf{0.136} \\
+ \pname\ [ours] & \textbf{18.36} & \textbf{0.324} & \textbf{0.377}  & \textbf{25.39} & 0.774 & 0.149 \\

\bottomrule
\end{tabular}
\vspace{0.2cm}
\caption{Results on the \textit{bicycle} scene of the mip-NeRF 360 dataset~\cite{barron2022mip360}, for sparser and denser settings. The proposed \pname\ is applied on Instant-NGP~\cite{muller2022instant}, PyNeRF~\cite{turki2024pynerf}, and 3DGS~\cite{kerbl2023gaussiansplat}. Uncertainty estimates are unavailable for 3DGS.}
\label{tab:bicycle}
%\vspace{-0.2cm}
\end{center}
\end{table*}

%% file: tables/by_scene/stump.tex
\begin{table*}
\begin{center}
\begin{tabular}{l|ccc|ccc}
\toprule
& \multicolumn{3}{c|}{\textit{stump} - sparser} & \multicolumn{3}{c}{\textit{stump} - denser} \\
%Method & PSNR $\uparrow$ & SSIM $\uparrow$ & LPIPS $\downarrow$ \\
Method & PSNR & SSIM & LPIPS & PSNR & SSIM & LPIPS \\
\midrule

PyNeRF~\cite{turki2024pynerf} & 22.42 & 0.632 & 0.218 & 26.47 & 0.810 & 0.117 \\
+ \pname\ [ours] & \textbf{23.75} & \textbf{0.672} & \textbf{0.188} & \textbf{26.51} & \textbf{0.814} & \textbf{0.112} \\
\midrule
3DGS~\cite{kerbl2023gaussiansplat} & 19.72 & 0.409 & 0.311 & 23.81 & 0.666 & 0.177\\
+ \pname\ [ours] & \textbf{19.97} & \textbf{0.433} & \textbf{0.306} & \textbf{24.32} & \textbf{0.691} & \textbf{0.172}\\

\bottomrule
\end{tabular}
\vspace{0.2cm}
\caption{Results on the \textit{stump} scene of the mip-NeRF 360 dataset~\cite{barron2022mip360}, for sparser and denser settings. The proposed \pname\ is applied on PyNeRF~\cite{turki2024pynerf} and 3DGS~\cite{kerbl2023gaussiansplat}. Instant-NGP~\cite{muller2022instant} could not deal with this scene in the sparser setting (PSNR of 15.55), so it was not applied here. Uncertainty estimates are unavailable for 3DGS.}
\label{tab:stump}
%\vspace{-0.2cm}
\end{center}
\end{table*}

%% file: tables/by_scene/garden.tex
\begin{table*}
\begin{center}
\begin{tabular}{l|ccc|ccc}
\toprule
& \multicolumn{3}{c|}{\textit{garden} - sparser} & \multicolumn{3}{c}{\textit{garden} - denser} \\
%Method & PSNR $\uparrow$ & SSIM $\uparrow$ & LPIPS $\downarrow$ \\
Method & PSNR & SSIM & LPIPS & PSNR & SSIM & LPIPS \\
\midrule

Instant-NGP~\cite{muller2022instant} & 23.03 & 0.733 & 0.163 & 25.54 & 0.829 & 0.111 \\
+ \pname\ [ours] & \textbf{23.27} & \textbf{0.743} & \textbf{0.149} & \textbf{25.64} & \textbf{0.831} & \textbf{0.108} \\
\midrule
PyNeRF~\cite{turki2024pynerf} & 25.80 & 0.841 & 0.073 & \textbf{28.26} & \textbf{0.900} & \textbf{0.052} \\
+ \pname\ [ours] & \textbf{26.18} & \textbf{0.845} & \textbf{0.071} & 27.88 & 0.897 & 0.055 \\
\midrule
3DGS~\cite{kerbl2023gaussiansplat} & 21.97 & 0.670 & 0.140 & 27.44 & 0.889 & 0.063 \\
+ \pname\ [ours] & \textbf{23.18} & \textbf{0.730} & \textbf{0.106} & \textbf{27.93} & \textbf{0.900} & \textbf{0.048} \\

\bottomrule
\end{tabular}
\vspace{0.2cm}
\caption{Results on the \textit{garden} scene of the mip-NeRF 360 dataset~\cite{barron2022mip360}, for sparser and denser settings. The proposed \pname\ is applied on Instant-NGP~\cite{muller2022instant}, PyNeRF~\cite{turki2024pynerf}, and 3DGS~\cite{kerbl2023gaussiansplat}. Uncertainty estimates are unavailable for 3DGS.}
\label{tab:garden}
%\vspace{-0.2cm}
\end{center}
\end{table*}

%% file: tables/by_scene/bonsai.tex
\begin{table*}
\begin{center}
\begin{tabular}{l|ccc|ccc}
\toprule
& \multicolumn{3}{c|}{\textit{bonsai} - sparser} & \multicolumn{3}{c}{\textit{bonsai} - denser} \\
%Method & PSNR $\uparrow$ & SSIM $\uparrow$ & LPIPS $\downarrow$ \\
Method & PSNR & SSIM & LPIPS & PSNR & SSIM & LPIPS \\
\midrule

Instant-NGP~\cite{muller2022instant} & 19.41 & 0.699 & 0.335 & \textbf{26.96} & \textbf{0.894} & \textbf{0.098} \\
+ \pname\ [ours] & \textbf{19.78} & \textbf{0.711} & \textbf{0.310} & 26.77 & 0.892 & 0.100 \\
\midrule
PyNeRF~\cite{turki2024pynerf} & 23.61 & 0.837 & 0.137 & \textbf{29.40} & \textbf{0.932} & \textbf{0.063} \\
+ \pname\ [ours] & \textbf{24.70} & \textbf{0.854} & \textbf{0.118} & 29.37 & 0.931 & \textbf{0.063} \\
\midrule
3DGS~\cite{kerbl2023gaussiansplat} & \textbf{23.32} & \textbf{0.743} & \textbf{0.212} & \textbf{30.88} & \textbf{0.938} & \textbf{0.078} \\
+ \pname\ [ours] & 23.09 & 0.705 & 0.217 & 30.61 & 0.931 & 0.087 \\

\bottomrule
\end{tabular}
\vspace{0.2cm}
\caption{Results on the \textit{bonsai} scene of the mip-NeRF 360 dataset~\cite{barron2022mip360}, for sparser and denser settings. The proposed \pname\ is applied on Instant-NGP~\cite{muller2022instant}, PyNeRF~\cite{turki2024pynerf}, and 3DGS~\cite{kerbl2023gaussiansplat}. Uncertainty estimates are unavailable for 3DGS.}
\label{tab:bonsai}
%\vspace{-0.2cm}
\end{center}
\end{table*}

%% file: tables/by_scene/kitchen.tex
\begin{table*}
\begin{center}
\begin{tabular}{l|ccc|ccc}
\toprule
& \multicolumn{3}{c|}{\textit{kitchen} - sparser} & \multicolumn{3}{c}{\textit{kitchen} - denser} \\
%Method & PSNR $\uparrow$ & SSIM $\uparrow$ & LPIPS $\downarrow$ \\
Method & PSNR & SSIM & LPIPS & PSNR & SSIM & LPIPS \\
\midrule

Instant-NGP~\cite{muller2022instant} & 20.75 & 0.709 & 0.232 & \textbf{26.28} & 0.837 & \textbf{0.135} \\
+ \pname\ [ours] & \textbf{21.70} & \textbf{0.753} & \textbf{0.195} & 25.97 & \textbf{0.839} & 0.136 \\
\midrule
PyNeRF~\cite{turki2024pynerf} & 23.16 & 0.840 & 0.119 & \textbf{28.77} & \textbf{0.919} & \textbf{0.069} \\
+ \pname\ [ours] & \textbf{24.08} & \textbf{0.872} & \textbf{0.099} & 28.62 & 0.916 & 0.073 \\
\midrule
3DGS~\cite{kerbl2023gaussiansplat} & 21.95 & 0.770 & 0.149 & 30.13 & \textbf{0.928} & \textbf{0.062} \\
+ \pname\ [ours] & \textbf{22.56} & \textbf{0.797} & \textbf{0.130} & \textbf{30.21} & \textbf{0.928} & 0.064 \\

\bottomrule
\end{tabular}
\vspace{0.2cm}
\caption{Results on the \textit{kitchen} scene of the mip-NeRF 360 dataset~\cite{barron2022mip360}, for sparser and denser settings. The proposed \pname\ is applied on Instant-NGP~\cite{muller2022instant}, PyNeRF~\cite{turki2024pynerf}, and 3DGS~\cite{kerbl2023gaussiansplat}. Uncertainty estimates are unavailable for 3DGS.}
\label{tab:kitchen}
%\vspace{-0.2cm}
\end{center}
\end{table*}

%% file: tables/by_scene/room.tex
\begin{table*}
\begin{center}
\begin{tabular}{l|ccc|ccc}
\toprule
& \multicolumn{3}{c|}{\textit{room} - sparser} & \multicolumn{3}{c}{\textit{room} - denser} \\
%Method & PSNR $\uparrow$ & SSIM $\uparrow$ & LPIPS $\downarrow$ \\
Method & PSNR & SSIM & LPIPS & PSNR & SSIM & LPIPS \\
\midrule

Instant-NGP~\cite{muller2022instant} & 22.68 & 0.778 & 0.248 & 27.62 & 0.883 & 0.143 \\
+ \pname\ [ours] & \textbf{24.14} & \textbf{0.801} & \textbf{0.217} & \textbf{28.07} & \textbf{0.886} & \textbf{0.139} \\
\midrule
PyNeRF~\cite{turki2024pynerf} & 23.85 & 0.825 & 0.168 & 27.24 & \textbf{0.913} & 0.092 \\
+ \pname\ [ours] & \textbf{24.81} & \textbf{0.841} & \textbf{0.146} & \textbf{27.28} & 0.911 & \textbf{0.091} \\
\midrule
3DGS~\cite{kerbl2023gaussiansplat} & 19.67 & 0.677 & 0.301 & 29.50 & 0.903 & 0.129 \\
+ \pname\ [ours] & \textbf{20.89} & \textbf{0.715} & \textbf{0.271} & \textbf{30.02} & \textbf{0.907} & \textbf{0.128} \\

\bottomrule
\end{tabular}
\vspace{0.2cm}
\caption{Results on the \textit{room} scene of the mip-NeRF 360 dataset~\cite{barron2022mip360}, for sparser and denser settings. The proposed \pname\ is applied on Instant-NGP~\cite{muller2022instant}, PyNeRF~\cite{turki2024pynerf}, and 3DGS~\cite{kerbl2023gaussiansplat}. Uncertainty estimates are unavailable for 3DGS.}
\label{tab:room}
%\vspace{-0.2cm}
\end{center}
\end{table*}